\documentclass[twocolumn]{article}

\usepackage[colorlinks,urlcolor=blue,linkcolor=blue,citecolor=blue]{hyperref}
\usepackage{color,array}
\usepackage{graphicx}

\usepackage{amsthm}
\newtheorem{theorem}{Theorem}

\usepackage{todonotes}
\usepackage{soul}  
\usepackage{comment}
\usepackage{amsmath}
\usepackage{colortbl}  
\usepackage{standalone}
\usepackage{pdflscape}  
\usepackage{fp}  

\usepackage{tikz}
\usetikzlibrary{positioning,fit,patterns,matrix}
\usepackage{tkz-euclide}
\usepackage{pgfplots}
\pgfplotsset{compat=1.18}
\usepgfplotslibrary{groupplots}
\usepackage{subcaption}  
\usepackage{bbm}  
\usepackage{amssymb}  
\usepackage{multirow}

\begin{document}

\title{Unimodal Distributions for Ordinal Regression}

\author{Jaime S. Cardoso, Ricardo P. M. Cruz, and Tomé Albuquerque\\
\small Faculty of Engineering, University of Porto, Portugal\\
\small INESC TEC, Portugal\\
\small (corresponding author: jaime.cardoso@fe.up.pt)\\
\small (published version: \href{https://doi.org/10.1109/TAI.2025.3549740}{doi:10.1109/TAI.2025.3549740})}

\date{}

\maketitle

\begin{abstract}  
In many real-world prediction tasks, the class labels contain information about the relative order between the labels that are not captured by commonly used loss functions such as multicategory cross-entropy. In ordinal regression, many works have incorporated ordinality into models and loss functions by promoting unimodality of the probability output. However, current approaches are based on heuristics, particularly non-parametric ones, which are still insufficiently explored in the literature.
We analyze the set of unimodal distributions in the probability simplex, establishing fundamental properties and giving new perspectives to understand the ordinal regression problem. Two contributions are then proposed to incorporate the preference for unimodal distributions into the predictive model: 1)~UnimodalNet, a new architecture that by construction ensures the output is a unimodal distribution, and 2)~Wasserstein Regularization, a new loss term that relies on the notion of projection in a set to promote unimodality. Experiments show that the new architecture achieves top performance, while the proposed new loss term is very competitive while maintaining high unimodality.
\end{abstract}




\section{Introduction}\label{sec:introduction}

Ordinal regression (sometimes also called ordinal classification) is applied to data where the features of the n-th example $\mathbf{x}_n \in {\cal X}$ correspond to a label from a set of elements ${\cal C} := \{ c_1,\dots , c_K\}$ that have a well-defined ranking or order $c_1 \prec c_2 \prec \dots \prec c_K$.  However, unlike traditional metric regression, quantitative differences or distances cannot be assumed to exist between classes. The goal is to find a reliable rule or regression function $h\colon  {\cal X} \rightarrow  {\cal C}$ from the domain of features ${\cal X}$ to the domain of ordinal labels ${\cal C}$.

Applications of ordinal regression include age estimation~\cite{FGNET}, cancer grading~\cite{ordinal_albuquerque2021}, photographs dating~\cite{HCI}, diabetic retinopathy grading~\cite{Beckham2017}, Alzheimer's disease progression~\cite{Wang2024}, time series on temperature~\cite{gomez2023one}, survival analysis~\cite{perez2017fine,rivera2023ordinal}, facial expression intensity estimation~\cite{Xu2024}, and evaluating the quality of manufacturing goods~\cite{vargas2023deep}.

In the published literature for applied problems, it is still common to ignore the order of the labels and apply categorical algorithms to such data, which often leads to the application of categorical cross-entropy loss (CE) in neural networks. Problematically, categorical loss assumes that all mislabelling by $h$ is equally wrong. Although the problem with this practice has been known for more than 40 years~\cite{POM}, it is still common to implicitly or explicitly assume that ordinal data or labels exist on an interval or ratio scale~\cite{Ashraf2022}. Interpretability methods (xAI) have been shown to produce better explanations when ordinal methods are used~\cite{barbero2023evaluating}, which suggests these methods tend to produce better models.

\begin{figure}
\centering
\begin{tikzpicture}[font=\small]
\node (fgnet) {\includegraphics[width=1.5em]{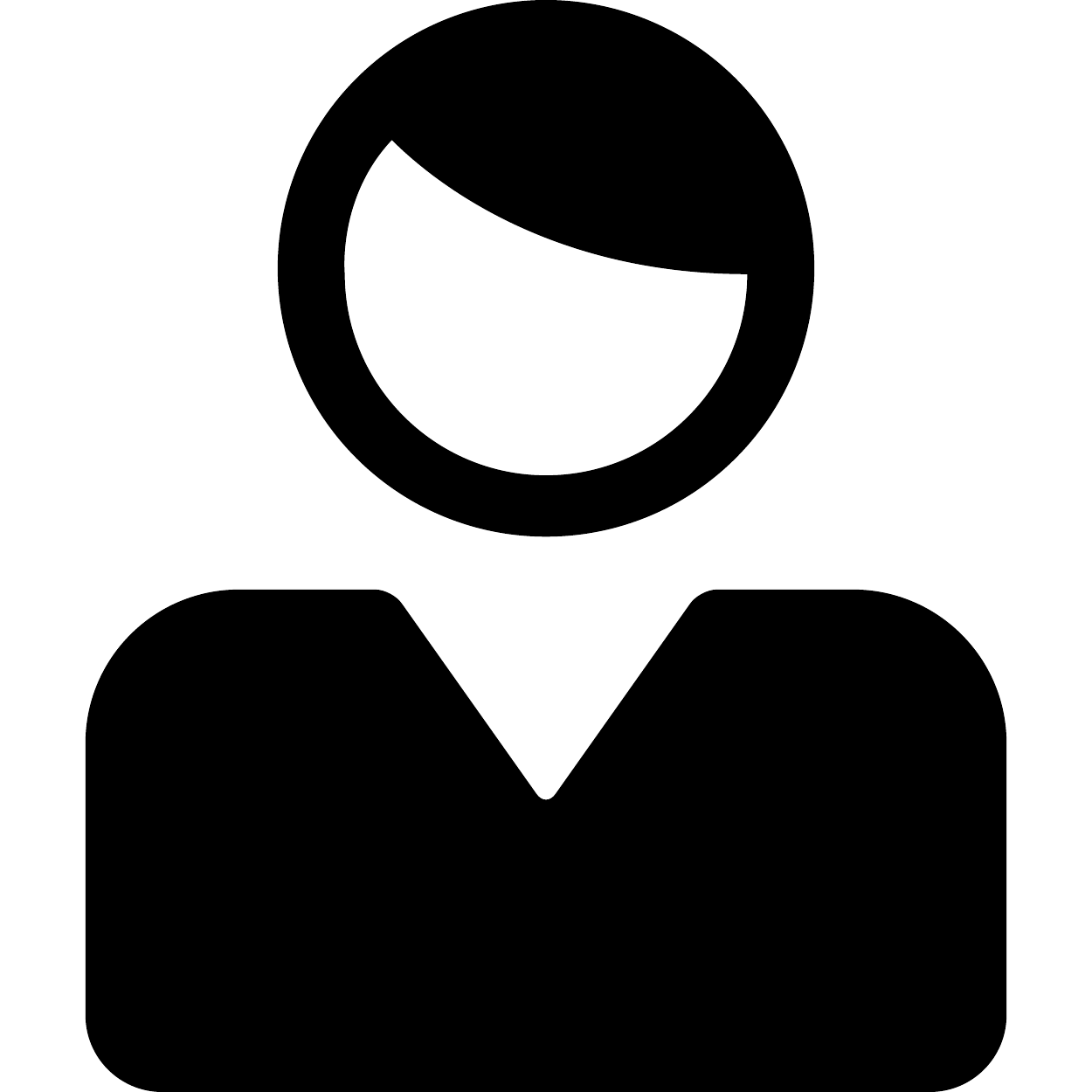} FGNET Age Estimation Dataset};
\draw[-latex] ([yshift=2ex]fgnet.south west) -| (-3,-0.5);
\draw[-latex] ([yshift=2ex]fgnet.south east) -| (+3,-0.5);
\end{tikzpicture}

\subfloat[Multimodal output distribution.\label{fig:multimodalplot}]{%
\begin{tikzpicture}[font=\small]
\begin{axis}[xbar, width=9em, height=30ex, xmin=0, xmax=0.6,
ytick={1,2,3,4,5,6},
ticklabel style = {font=\footnotesize},
yticklabels={$\text P(\text{Age}{=}10)$, $\text P(\text{Age}{=}20)$, $\text P(\text{Age}{=}30)$, $\text P(\text{Age}{=}40)$, $\text P(\text{Age}{=}50)$, $\text P(\text{Age}{=}60)$}
]
\addplot [bar shift=0pt, pattern=north east lines, pattern color=gray] coordinates {(0.05, 2) (0.05, 4) (0.05, 5)};
\addplot [bar shift=0pt, red, pattern=north east lines, pattern color=red] coordinates {(0.1, 1) (0.4, 3) (0.3, 6)};
\end{axis}
\end{tikzpicture}}\hfill%
\subfloat[Unimodal output distribution.]{%
\begin{tikzpicture}[font=\small]
\begin{axis}[xbar, width=9em, height=30ex, xmin=0, xmax=0.6,
ytick={1,2,3,4,5,6},
ticklabel style = {font=\footnotesize},
yticklabels={$\text P(\text{Age}{=}10)$, $\text P(\text{Age}{=}20)$, $\text P(\text{Age}{=}30)$, $\text P(\text{Age}{=}40)$, $\text P(\text{Age}{=}50)$, $\text P(\text{Age}{=}60)$}
]
\addplot [bar shift=0pt, pattern=north east lines, pattern color=gray] coordinates {(0.05, 1) (0.2, 2) (0.2, 4) (0.125, 5) (0.075, 6)};
\addplot [bar shift=0pt, red, pattern=north east lines, pattern color=red] coordinates {(0.3, 3)};
\end{axis}
\end{tikzpicture}}

\caption{Example of possible output probability distributions. Even if both outputs agree on the majority class, only the unimodal distribution is consistent with an ordinal regression task.}%
\label{fig:exampleelastic}%
\end{figure}

Among methods recently developed specifically for ordinal regression, unimodal constraints on the probability distribution of the output have proven to be strong contenders. For a class $c_\ell$ for which the model outputs the highest posterior probability, $\ell=\arg_k\max \text P(y=c_k\,|\,\mathbf x)$, it would be expected for the second-highest posterior probability to be either $\text P(y=c_{\ell-1}\,|\,\mathbf x)$ or $\text P(y=c_{\ell+1}\,|\,\mathbf x)$, given the previously defined class order, $c_{\ell-1}\prec c_\ell\prec c_{\ell+1}$. The previous reasoning also applies to the ranking order from the third, fourth, et cetera posterior probabilities. For example, it makes no sense for a model to produce a high probability for ``Hot'' and for ``Cold'', but a small probability for ``Warm''. Furthermore, it is common for ordinal regression metrics to penalize errors more strongly when the distance between the predicted and the true class is farther apart than when it is closer. Some examples of output distributions are provided by Fig.~\ref{fig:exampleelastic}; only the unimodal case is consistent with ordinal regression. However, the current ordinal literature does not necessarily promote unimodality and can produce inconsistencies like those in Fig.~\ref{fig:multimodalplot}. The papers that do promote unimodality are either based on heuristics (such as CO~\cite{ordinal_albuquerque2021,math10060980}) or assume that predictions must follow a parametric probability distribution (such as the Binomial~\cite{Costa2005Classification}), which is often an unrealistic assumption that hinders performance.

The paper highlights the limitations of current approaches that rely on unimodal distributions~\cite{Costa2005Classification,Beckham2017,Belharbi2019DeepOC,ordinal_albuquerque2021} and presents three major contributions:
\begin{enumerate}
\item An analysis is performed on the properties of unimodal distributions. This analysis shows that the set of unimodal distributions is contiguous, making it reasonable to navigate through this space using iterative optimization methods to find the best model. Two novel and independent solutions are then presented.
\item UnimodalNet: A neural network architecture is proposed that, by construction, imposes hard constraints to restrict the probability output to the region where the unimodal subset is located.
\item Wasserstein Regularization: A regularization loss term is proposed that projects the probability output to the closest unimodal distribution and penalizes deviations from that distribution.
\end{enumerate}



\section{Related work}

Assume that in a classification task, the instances have one of $K$ classes, whose labels are $c_1$ to $c_K$, which correspond to the natural order of the ordinal classes.

Typically, a neural network is trained to perform multi-class classification by minimizing the cross-entropy loss for the entire training set,
\begin{equation}
\label{eq:ce}
\text{CE}({\mathbf y}_n,\mathbf{\hat y}_n)=-\sum_{k=1}^Ky_{nk}\log(\hat y_{nk}),
\end{equation}
where, for the $n$-th observation, ${\mathbf y}_n = [y_{n 1}\dots y_{nk} \dots y_{nK}]$, with $y_{nk} \in \{0,1\}$, represents the respective one-hot encoding of the corresponding label $c_n \in \cal{C}$ and $\mathbf{\hat y}_n = [\hat y_{n 1}\dots \hat y_{nk} \dots \hat y_{nK}]$, with $\hat y_{nk} \in [0,1]$, being the respective vector of output probabilities assigned by the model for that $n$-th observation. 
Naturally, $\sum_{k=1}^K y_{nk} = \sum_{k=1}^K \hat y_{nk} = 1$, so ${\mathbf y}_n$ and $\mathbf{\hat y}_n$ are points in the $(K-1)$ dimensional probability simplex (each point in the K-1 dimensional probability simplex represents a probability distribution over K mutually exclusive events).

However, CE has limitations when applied to ordinal data. 
Defining $k_n^\star\in \{1,\dots,K\}$ as the index or rank of the true class of observation ${\mathbf x}_n$ (the position where $y_{nk} = 1$ in the one hot encoding vector ${\mathbf y}_n$), it is then clear that 
\begin{equation}
\text{CE}({\mathbf y}_n,\mathbf{\hat y}_n)=-\log(\hat y_{nk_n^\star}).    
\end{equation}

\subsection{Beyond cross-entropy}

Intuitively, CE tries to maximize the probability in the output corresponding to the true class, ignoring all the other probabilities. For this loss, an error between classes $c_1$ and $c_2$ is treated as the same as an error between $c_1$ and $c_K$, which is undesirable for ordinal problems. 

Furthermore, the loss does not constrain the model to produce unimodal probabilities, so inconsistencies can be produced, such as $\hat y_{nj}>\hat y_{n\ell}<\hat y_{ni}$, even when $1\leq j < \ell < i\leq K$. This inconsistency can be verified in the examples provided in the results section (Fig.~\ref{fig:outputs}).

Cross-entropy is a fair approach for nominal data, where no additional information is available. By concentrating just on the mode of the distribution and disregarding all other values in the output probability vector, the ordinal information inherent in the data is ignored. 
However, for ordinal data, the order can be explored to further regularize learning.

Ordinal regression methods include a variety of techniques, which can be organized using different rationales. Guti\'errez et al.~\cite{7161338} proposed a taxonomy of ordinal regression methods: (i)~binary decomposition approaches (decompose the ordinal problem into several binary ones, which are separately solved by multiple models or by one multiple-output model), (ii)~na\"{\i}ve approaches (the model is obtained by using other standard machine learning prediction algorithms), and (iii)~threshold methods (based on the general idea of approximating a real value predictor and then dividing the real line into intervals).

Focusing only on probability-based approaches, cumulative link models were one of the first approaches to emerge~\cite{POM,Frank2001simple}. The Proportional Odds Model (POM)~\cite{POM} proposes learning $\text P(y\leq c_k\mid\mathbf x)=\sigma(\theta_k-w^\intercal\mathbf x)$, for each $1\leq k<K$. Probabilities are, therefore, encoded as a cumulative distribution, which can naturally be posteriorly converted to a mass function. In POM, weights are common to all classes, and each class $k$ learns a single parameter $\theta_k$. Ordinal Encoding (OE)~\cite{Frank2001simple,cheng2008neural} proposes learning $\text P(y>c_k)$ with different weights for each class, and, during inference, the predicted class is then $\ell=\sum_{k=1}^{K-1}\mathbbm1(\text P(y>c_k))$, where $\mathbbm1(\cdot)$ is the indicator function. Another method, CDW-CE, promotes ordinality by combining CE with MAE (mean absolute error) in the loss function~\cite{polat2022class}.

More recently, several works have promoted or forced the output distribution to be unimodal. This can be achieved either by the construction of the model or by the use of an appropriate loss during training.

\paragraph{Parametric models}
One approach involves restricting the output of the model to that of a discrete probability distribution, such as the Binomial or Poisson. For the Binomial distribution~\cite{Costa2005Classification}, $\mathcal B(n,p)$ the support of the distribution is known, $n=K{-}1$, and the only parameter left to be estimated is the shape of the distribution, $p$. The model outputs a single output, the shape of the distribution, which is then converted into posterior probabilities using the Binomial probability mass function, $\text P(y=c_k\,|\,\mathbf x)=\binom{n}{k-1}p^{k-1}(1-p)^{n-k-1}$, for each $k\in\{1,\dots, K\}$, where $p$ is the output produced by the model. This may be seen as an activation function. A common loss, such as cross-entropy, may then be used to train the model.

Similarly, the Poisson probability mass function may be enforced. Furthermore, \cite{Beckham2017} also proposes to control the variance of the distribution through a learnable softmax temperature term ($\tau$). In \cite{ARAUJO2020101715}, the authors propose a diabetic retinopathy grading CAD system that provides an estimation of the uncertainty of the decision by imposing a Gaussian distribution centered on the predicted class.

\paragraph{Parametric losses}
Such parametric distributions (Binomial and Poisson) have also been used as an alternative to one-hot encoding for the target label distribution. Instead of a loss that penalizes only deviations from the true class, Unimodal Regularization (UR)~\cite{liu2020unimodal} modifies \eqref{eq:ce} so that $y_{nk}$ follows a parametric distribution (e.g., instead of $\mathbf y=[0,0,1,0]$ for $c_n=3$, UR penalizes deviations from $\mathbf y=[0.02, 0.14, 0.46, 0.38]$).

\paragraph{Non-parametric models}
Instead of assuming an a-priori probability distribution, ORD-ACL and VS-SL~\cite{yamasaki2022unimodal} have been proposed to ensure unimodality when converting the logits to probabilities. Both of the models start by applying an ORD transformation to ensure that logits $z$ are ascending $z'_k=z_1+\sum_{\ell=2}^k\rho(z_k)$ with $\rho\colon\mathbb R\to[0,\infty)$ (e.g., $\rho(u)=u^2$ or $\rho(u)=\exp(u)$). The two models then use different link functions: in ORD-ACL, logits $z\in\mathbb R^{K-1}$ and the link presupposed what is being modeled are adjacent probabilities, $\frac{\text P(Y=k\mid x)}{\text P(Y\in\{k,k+1\}\mid x)}$, while VS-SL uses a softmax but first transforms logits to have an inverted V-shape, $z''_k=-\tau(z'_k)$ where $\tau$ is a symmetric function like $\tau(u)=|u|$ or $\tau(u)=u^2$.

Both ORD-ACL and VS-SL are shown by the authors to produce a unimodal probability distribution. Variants are also proposed: the PO variant uses the aforementioned POM model~\cite{POM} as the learner; the HO variant is based on \cite{Beckham2017} and extends the models by incorporating a scaling factor that is learned for each observation, $s(x)$, so the scale of the probability distribution varies based on the input to handle overall heteroscedasticity.

\paragraph{Non-parametric losses}
CO2 (CE+O2) is a non-parametric regularization term called O2 that is added to Cross Entropy (CE) or another loss~\cite{Belharbi2019DeepOC,ordinal_albuquerque2021},
\begin{equation}
\text{O2}(y,\hat p)=\sum_{k=1}^y\max(0,\hat p_k-\hat p_{k+1})+\sum_{k=y}^K\max(0,\hat p_{k+1}-\hat p_k).
\end{equation}
The goal is to promote the neural network's output probabilities to follow a unimodal distribution. This is done by imposing a set of different constraints over all pairs of consecutive labels, which allows for a more flexible decision boundary relative to parametric approaches. These approaches have been generalized for quasi-unimodal distributions~\cite{math10060980}.

Fig.~\ref{figTaxonomy} summarizes the current unimodal approaches concerning the soft/hard constraint and parametric/non-parametric priority axes. We also highlight the two proposed model families, to be presented next.

\begin{figure}
\centering
\begin{tikzpicture}[thick, font=\small]
\matrix (mymatrix) [
matrix of nodes,
column 1/.style={nodes={rotate=90, anchor=center}},
column 2/.style={nodes={text width=11em, anchor=center}},
column 3/.style={nodes={text width=11em, anchor=center}},
row 3/.style={nodes={align=center}},
] {
non-parametric & c)\newline$\bullet$ ORD-ACL, VS-SL~\cite{yamasaki2022unimodal}\newline$\bullet$ \textbf{Contribution 1:} UnimodalNet & d)\newline$\bullet$ Non-parametric heuristics \cite{ordinal_albuquerque2021,math10060980}\newline$\bullet$ \textbf{Contribution 2}: Wasserstein Unimodal \\
parametric & a)\newline$\bullet$ Binomial Unimodal \cite{Costa2005Classification}\newline$\bullet$ Poisson Unimodal \cite{Beckham2017} & b)\newline $\bullet$ Unimodal Regularization (UR)~\cite{liu2020unimodal} \\
& hard unimodal (model) & soft unimodal (loss) \\
};

\draw[-latex, thick] (mymatrix-2-1.south west) -- (mymatrix-1-1.south east);
\draw[-latex, thick] (mymatrix-3-2.north west) -- (mymatrix-3-3.north east);

\draw[dashed] ([yshift=4.4em]mymatrix-3-2.north west) -- ([yshift=4.4em]mymatrix-3-3.north east);
\draw[dashed] ([xshift=-0.3em, yshift=10.5em]mymatrix-3-2.north east) -- ([xshift=-0.3em]mymatrix-3-2.north east);
\end{tikzpicture}
\caption{Summary of the current unimodal approaches, where the axes represent the soft/hard constraint and parametric/nonparametric priorities. The proposed contributions are also mentioned in the right families.}
\label{figTaxonomy}
\end{figure}

\section{Task definition}

Before delving into the proposal section, where the two main approaches proposed in this paper are described, we start by contributing with a theoretical analysis of the subset of unimodal distributions in the $(K-1)$ dimensional probability simplex.

\subsection{Understanding the set of unimodal distributions}
Let $\mathbf{p} = [p_1, \dots, p_K]$ be a distribution in the $K - 1$ dimensional probability simplex. We say $\mathbf{p}$ is unimodal with mode at $k^\star$ if $p_1\leq p_2\leq \dots \leq p_{k^\star}\geq p_{k^\star+1} \geq \dots \geq p_K$.
Note that the uniform distribution is considered unimodal (for any $k^\star$).

\begin{figure}
\centering
\newcommand{\incenterb}[4]{%
    \begin{tikzpicture}[font=\small]
        \tkzInit[xmax=7.7,ymax=6.5]
        \tkzClip
        \tkzDefPoint(2,1){#2} 
        \tkzDefPoint(4,5){#3} 
        \tkzDefPoint(6,1){#4} 
        \tkzDrawPolygon[color=black](#2,#3,#4)
        \tkzInCenter(#2,#3,#4)
        \tkzGetPoint{G}
        \node[below] at (G) {#1};

        \draw [-stealth](2,1) -- (1,0.4);
        \draw [-stealth](6,1) -- (7,0.4);
        \draw [-stealth](4,5) -- (4,6);
        
        \draw [dashed](G) -- (2,1);
        \draw [dashed](G) -- (6,1);

        \draw (2,1) -- (2.95,2.9) -- (G) -- (5.05,2.9) -- (6,1) -- cycle;
        
        \draw (2.95,2.9) -- (G) -- (5.05,2.9) -- (4,5) -- cycle;
        
        \draw[fill=black!20] (2,1) -- (G) -- (6,1) -- cycle;
        
        \node at (0.8,0.3) {$p_1$};
        \node at (7.2,0.3) {$p_3$};
        \node at (4,6.2) {$p_2$};
        
        \node at (1.4,1.1) {(1,0,0)};
        \node at (6.6,1.1) {(0,0,1)};
        \node at (4.6,5.1) {(0,1,0)};
        
        \node at (4,1.4) {Non Unimodal};
        
        \node[rotate=-147.5] at (3,2) {\rotatebox{180}{Mode $p_1$}};
        
        \node[rotate=-121] at (5,2) {\rotatebox{90}{Mode $p_3$}};
        
        \node[rotate=-90] at (4,3.5) {\rotatebox{180}{Mode $p_2$}};
        
        \filldraw [fill=white, draw=black] (5.9,5) rectangle (6.2,5.5);
        \filldraw [fill=white, draw=black] (6.2,5) rectangle (6.5,6);
        \filldraw [fill=white, draw=black] (6.5,5) rectangle (6.8,5.8);
        \draw (5.5,5) -- (7.2,5);
        \draw[-latex] (4.5, 3.5) to[bend right=40] (6.4, 4.9);

        \filldraw [fill=white, draw=black] (6.5,2.8) rectangle (6.8,3.3);
        \filldraw [fill=white, draw=black] (6.8,2.8) rectangle (7.1,3.6);
        \filldraw [fill=white, draw=black] (7.1,2.8) rectangle (7.4,3.8);
        \draw (6.1,2.8) -- (7.8,2.8);
        \draw[-latex] (5.2, 2.2) to[bend right=40] (7, 2.7);
        
        \filldraw [fill=white, draw=black] (0.5,1.8) rectangle (0.8,2.8);
        \filldraw [fill=white, draw=black] (0.8,1.8) rectangle (1.1,2.3);
        \filldraw [fill=white, draw=black] (1.1,1.8) rectangle (1.4,2.6);
        \draw (0.1,1.8) -- (1.8,1.8);
        \draw[-latex] (3.3, 1.25) to[bend left=20] (0.9, 1.7);

        \filldraw [fill=white, draw=black] (1.1,4) rectangle (1.4,5);
        \filldraw [fill=white, draw=black] (1.4,4) rectangle (1.7,4.8);
        \filldraw [fill=white, draw=black] (1.7,4) rectangle (2,4.5);
        \draw (0.7,4) -- (2.4,4);
        \draw[-latex] (2.8, 2.2) to[bend left=40] (1.6, 3.9);

    \end{tikzpicture}
}%
\incenterb{}{a}{b}{c}
\caption{Pictorial representation of unimodal and non-unimodal distributions in the 2D simplex.}
\label{simplexGraph}
\end{figure}   


\begin{theorem}
In the $(K - 1)$ dimensional probability simplex, the set of unimodal distributions with a fixed mode is a connected set. 
\end{theorem}
\begin{proof}
Let $\mathbf{p}=[p_1, \ p_2, \dots, p_{k^\star},\dots, p_K ]$ be a unimodal distribution with mode $k^\star$. 

Consider
\begin{equation}
\mathbf{q}(\delta)=\big[p_1-\delta p_1,\ p_2+\delta\frac{p_1}{k^\star-1},\ \dots,\ p_{k^\star}+\delta\frac{p_1}{k^\star-1},\ p_{k^\star+1},\ \dots,\ p_K \big],
\end{equation}
with $0\leq \delta \leq 1$. Note that (a)~$\mathbf{q}(\delta)$ is a unimodal distribution with mode $k^\star$ since the order of the probability values is preserved, (b)~$\mathbf{q}(\delta)$ is a continuous function in $\delta$, (c)~$q_1(1) = 0$, (d)~there is a path in the probability simplex connecting $\mathbf{p}$ to $\mathbf{q}(1)$ over unimodal distributions only.

Repeating the process sequentially for $p_2, \dots p_{k^\star-1}, p_K, p_{K-1}, \dots,  p_{k^\star+1}$, one can continuously transform any unimodal distribution with mode in $k^\star$ in the distribution with $p_{k^\star} = 1$ and zero in all the other values. Therefore, the set of unimodal distributions with a fixed mode in $k^\star$ is connected.

One can further confirm that the set of all unimodal distributions is indeed connected. It suffices to note that the uniform distribution is in the set of unimodal distributions for any $k^\star$.
\end{proof}

Fig.~\ref{simplexGraph} illustrates the result for $K=3$.
In the context of this property, it is trivially proved that the set of unimodal distributions with mode $k^\star$ is convex, but the set of all unimodal distributions is not convex.

Finally, it is interesting to discuss the ``size'' of the set of unimodal distribution within the $(K - 1)$ dimensional probability simplex.
Let $us(K)$ be the fraction of the points in $(K-1)$ simplex that corresponds to unimodal distributions and $ns(K)$ the fraction of non-unimodal distributions.
When $K=3$, one-third of the distributions are not unimodal, corresponding to the distributions where $p_2$ is the smallest value of the three: $us(3)=2/3$ and $ns(K)=1/3$.

For general $K$, if $\min \{p_i\colon i=1, \dots, K\} \not \in  \{p_1,p_K\}$ then the distribution is not unimodal. By symmetry, it is clear that $ns(K) \geq (K-2)/K$.
It is possible to show that
\begin{equation}ns(K) = \frac{(K-2)}{K} + \frac{2}{K}ns(K-1).\end{equation}

\begin{theorem}
The fraction of points, $ns(K)$, in $(K-1)$ simplex that corresponds to non-unimodal distributions is given by 
$ns(K) = \frac{(K-2)}{K} + \frac{2}{K}ns(K-1)$
\end{theorem}
\begin{proof}
For $K=3$, the result is trivial, as already presented.
For $K>3$, it was already stated that if $\min \{p_i\colon i=1, \dots, K\} \not \in  \{p_1,p_K\}$ then the distribution is not unimodal. 
If $\min \{p_i\colon i=1, \dots, K\} = p_1$, consider the distribution $\mathbf{q} = [p_2/(1-p_1),\ p_3/(1-p_1),\ \dots,\ p_K/(1-p_1)]$ in the $(K-2)$ simplex. Then $\mathbf{p}$ is non-unimodal $\iff$ $\mathbf{q}$ is non-unimodal. And this happens for $ns(K-1)$ of the $\mathbf{q}$ distributions. A similar argument applies when $\min \{p_i\colon i=1, \dots, K\} = p_K$.
\end{proof}

Table~\ref{unimodalPercentage} presents the fraction of distributions that are (non-)unimodal, for several $K$ values.
\begin{table}
\centering
\caption{Fraction of Unimodal Distributions}
\label{unimodalPercentage}
\begin{tabular}{ |c|c|c|c|c| } 
\hline
$K$ & 3 & 4 & 5 & 6\\
\hline
unimodal     & 0.667 & 0.333 & 0.133 & 0.044 \\
non-unimodal & 0.333 & 0.667 & 0.867 & 0.956 \\
\hline
\end{tabular}
\end{table}
Since the subset of unimodal distributions is considerably smaller than the complete set of probability distributions, even for small $K$, there is clear potential in exploring this knowledge in the learning of predictive models for ordinal regression.

\section{Proposal}

The fact that the set of unimodal distributions is contiguous makes it reasonable to navigate in this space using iterative optimization methods to find the best model. The fact that the subset of unimodal distributions is only a small fraction of the complete distribution set means that it is worth exploring this prior during the learning phase. Next, we propose an architectural layer for a neural network that enforces unimodal distributions in the output. This hard nonparametric constraint can be especially important for small datasets.
We then propose a soft regularization alternative by adding a penalty term to the loss function that optimizes the model. The penalty term encourages the model to stay/return to the set of models that produce unimodal distributions (a violation of the constraint incurs a penalty in the loss function). This soft constraint may be preferable if ordinality is incompletely present in our task, perhaps due to inadequate data representation. Depending on the specific problem and the characteristics of the data, one or the other mechanism may be preferred.

\subsection{Non-parametric unimodal architecture}
\label{sec:hard-proposal}

Here, we present a non-parametric restriction, forcing unimodal distributions. Instead of penalizing deviations from unimodality, like CO2~\cite{ordinal_albuquerque2021} and our proposal in the next section, the model outputs are designed to always guarantee unimodality. An activation function is proposed that produces cumulative sums of the outputs of the neural network from left to right and right to left, thus yielding a monotonously increasing and a monotonously decreasing sequence, respectively. A unimodal distribution is then constructed from these two sequences. Fig.~\ref{fig:unimodalnet} provides an illustrative example of the procedure.

\begin{figure}%
\def\first{0.5}%
\def\second{2}%
\def\third{1}%
\def\fourth{2}%
\centering
\begin{subfigure}[t]{0.22\textwidth}
\centering
\begin{tikzpicture}[x=0.7cm, y=0.3cm]
\node at (0,0) {$y_{i1}$};
\node at (1,0) {$y_{i2}$};
\node at (2,0) {$y_{i3}$};
\node at (3,0) {$y_{i4}$};
\path[draw=blue, thick, fill=lightgray] (-0.5, 1) rectangle ++(1, \first);
\path[draw, fill=lightgray] (0.5, 1) rectangle ++(1, \second);
\path[draw, fill=lightgray] (1.5, 1) rectangle ++(1, \third);
\path[draw=red, thick, fill=lightgray] (2.5, 1) rectangle ++(1, \fourth);
\end{tikzpicture}
\caption{Neural network output.}
\end{subfigure}\hfill%
\begin{subfigure}[t]{0.22\textwidth}
\centering
\begin{tikzpicture}[x=0.7cm, y=0.3cm]
\node at (0,0) {$y^{(\ell r)}_{i1}$};
\node at (1,0) {$y^{(\ell r)}_{i2}$};
\node at (2,0) {$y^{(\ell r)}_{i3}$};
\node at (3,0) {$y^{(\ell r)}_{i4}$};
\path[draw=blue, thick, pattern=north east lines, pattern color=blue] (-0.5, 1) rectangle ++(1, \first);
\path[draw, pattern=north east lines, pattern color=blue] (0.5, 1) rectangle ++(1, \first+\second);
\path[draw, pattern=north east lines, pattern color=blue] (1.5, 1) rectangle ++(1, \first+\second+\third);
\path[draw, pattern=north east lines, pattern color=blue] (2.5, 1) rectangle ++(1, \first+\second+\third+\fourth);
\end{tikzpicture}
\caption{Left-right cumulative sum.}
\end{subfigure}\hfill%
\begin{subfigure}[t]{0.22\textwidth}
\centering
\begin{tikzpicture}[x=0.7cm, y=0.3cm]
\node at (0,0) {$y^{(r\ell)}_{i1}$};
\node at (1,0) {$y^{(r\ell)}_{i2}$};
\node at (2,0) {$y^{(r\ell)}_{i3}$};
\node at (3,0) {$y^{(r\ell)}_{i4}$};
\path[draw, pattern=north west lines, pattern color=red] (-0.5, 1) rectangle ++(1, \first+\second+\third+\fourth);
\path[draw, pattern=north west lines, pattern color=red] (0.5, 1) rectangle ++(1, \second+\third+\fourth);
\path[draw, pattern=north west lines, pattern color=red] (1.5, 1) rectangle ++(1, \third+\fourth);
\path[draw=red, thick, pattern=north west lines, pattern color=red] (2.5, 1) rectangle ++(1, \fourth);
\end{tikzpicture}
\caption{Right-left cumulative sum.}
\end{subfigure}\hfill%
\begin{subfigure}[t]{0.22\textwidth}
\centering
\begin{tikzpicture}[x=0.7cm, y=0.3cm]
\node at (0,0) {$y'_{i1}$};
\node at (1,0) {$y'_{i2}$};
\node at (2,0) {$y'_{i3}$};
\node at (3,0) {$y'_{i4}$};
\FPeval{\minfirst}{min(\first,\first+\second+\third+\fourth)}
\FPeval{\minsecond}{min(\first+\second,\second+\third+\fourth)}
\FPeval{\minthird}{min(\first+\second+\third,\third+\fourth)}
\FPeval{\minfourth}{min(\first+\second+\third+\fourth,\fourth)}
\path[draw, pattern=north east lines, pattern color=blue] (-0.5, 1) rectangle ++(1, \minfirst);
\path[draw, pattern=north east lines, pattern color=blue] (0.5, 1) rectangle ++(1, \minsecond);
\path[draw, pattern=north west lines, pattern color=red] (1.5, 1) rectangle ++(1, \minthird);
\path[draw, pattern=north west lines, pattern color=red] (2.5, 1) rectangle ++(1, \minfourth);
\end{tikzpicture}
\caption{Final output (element-wise minimum).}
\end{subfigure}
\caption{Exemplification of UnimodalNet.}
\label{fig:unimodalnet}
\end{figure}

Let ${\mathbf v} \in R^K$ be the output of a certain (deep NN) model.
Let $\mathbf{z} \in (R_0^+)^K$, with $z_i = f(v_i)$, being $f(.)$ a non-negative function (a non-negative function has a range of 0 to infinity). If $v_i$ is already always non-negative (${\mathbf v} \in(R_0^+)^K$~) by construction (for instance, and if the last activation unit is a ReLU or its smooth approximation Softplus), then one can simplify and make $\mathbf{z}={\mathbf v}$. 

\begin{figure}
\centering
\begin{tikzpicture}[node distance=0.5ex and 2em, font=\small]
\node (z1) {$v_1$\vphantom{$z^{\ell r}_1$}};
\node[below=of z1] (z2) {$v_2$\vphantom{$z^{\ell r}_1$}};
\node[below=of z2] (z3) {$v_3$\vphantom{$z^{\ell r}_1$}};

\node[right=of z1] (zz1) {$z_1$\vphantom{$z^{\ell r}_1$}};
\node[below=of zz1] (zz2) {$z_2$\vphantom{$z^{\ell r}_1$}};
\node[below=of zz2] (zz3) {$z_3$\vphantom{$z^{\ell r}_1$}};

\draw[-latex] (z1) -- node[above] {relu} (zz1);
\draw[-latex] (z2) -- node[above] {relu} (zz2);
\draw[-latex] (z3) -- node[above] {relu} (zz3);

\node[right=6em of zz1, minimum width=2.55em] (zm1) {$z^{\ell r}_1$};
\node[below=of zm1, minimum width=2.55em] (zm2) {$z^{\ell r}_2$};
\node[below=of zm2, minimum width=2.55em] (zm3) {$z^{\ell r}_3$};

\node[draw, inner sep=2pt, left=3em of zm2] (zz2+) {$+$};
\node[draw, inner sep=2pt, left=1em of zm3] (zz3+) {$+$};
\coordinate[left=0.2em of zz3+] (zz3+pre);


\draw[-latex] (zz1) -- (zm1);

\draw (zz2) -- (zz2+); \draw[-latex] (zz2+) -- (zm2);
\draw (zz3) -- (zz3+); \draw[-latex] (zz3+) -- (zm3);

\draw ([xshift=1.5em]zz1.east) |- ([yshift=0.8ex]zz2+.west);
\draw ([xshift=0.6em]zz2+.east) |- ([yshift=0.8ex]zz3+.west);


\node[below=of zm3, minimum width=2.55em] (zd3) {$z^{r \ell}_3$};
\node[below=of zd3, minimum width=2.55em] (zd2) {$z^{r \ell}_2$};
\node[below=of zd2, minimum width=2.55em] (zd1) {$z^{r \ell}_1$};

\node[black!60, left=3em of zd2, draw, inner sep=2pt] (zd2+) {$+$};
\node[black!60, left=1em of zd1, draw, inner sep=2pt] (zd1+) {$+$};

\draw[black!60, -latex] ([xshift=2.5em]zz3.east) |- (zd3);

\draw[black!60] ([xshift=1.7em]zz3.east) |- ([yshift=0.4ex]zd2+.west);
\draw[black!60] ([xshift=1.3em]zz2.east) |- ([yshift=-0.4ex]zd2+.west);
\draw[black!60, -latex] (zd2+) -- (zd2);

\draw[black!60] ([xshift=0.6em]zd2+.east) |- ([yshift=0.4ex]zd1+.west);
\draw[black!60] ([xshift=0.7em]zz1.east) |- ([yshift=-0.4ex]zd1+.west);
\draw[black!60, -latex] (zd1+) -- (zd1);


\node[draw, above right=-11ex and 6em of zd3, rotate=90, minimum width=9em] (softmax) {softmax};

\draw[-latex] (zm3.east) -- ++(2em,0) |- node[right, yshift=1.6ex] {min} ([yshift=1ex]softmax.north);
\draw[-latex] (zd3.east) -- ++(2em,0) |- ([yshift=1ex]softmax.north);

\draw[black!60, -latex] (zm2.east) -- ++(1.2em,0) |- node[right, xshift=0.8em, yshift=1.3ex] {min} ([yshift=-3.4ex]softmax.north);
\draw[black!60, -latex] (zd2.east) -- ++(1.2em,0) |- ([yshift=-3.4ex]softmax.north);

\draw[-latex] (zm1.east) -- ++(0.5em,0) |- node[right, xshift=1.5em, yshift=1.3ex] {min} ([yshift=-8ex]softmax.north);
\draw[-latex] (zd1.east) -- ++(0.5em,0) |- ([yshift=-8ex]softmax.north);


\node[right=3em of softmax.south, yshift=1ex] (y1) {$\hat y_1$};
\node[right=3em of softmax.south, yshift=-3.4ex] (y2) {$\hat y_2$};
\node[right=3em of softmax.south, yshift=-7.8ex] (y3) {$\hat y_3$};

\draw[latex-] (y1) -- ++(-3.8em,0);
\draw[latex-] (y2) -- ++(-3.8em,0);
\draw[latex-] (y3) -- ++(-3.8em,0);

\end{tikzpicture}
\caption{Illustration for three classes of the proposed activation function, which forces the model output to be a unimodal distribution.}
\label{unimodalNet}
\end{figure}   

Let $\mathbf{z}^{\ell r} \in (R_0^+)^K$ with 
\begin{eqnarray}
        z^{\ell r}_1 = z_1 \nonumber & \\
        z^{\ell r}_i = z^{\ell r}_{i-1} + z_{i}, & \quad i=2, \dots, K
\end{eqnarray}

Then $z^{\ell r}_i\geq z^{\ell r}_{i-1}$ by construction, for $i =2,\dots, K$.
Similarly, one can construct $\mathbf{z}^{r\ell}$, with $z^{r\ell}_i\leq z^{r\ell}_{i-1}$, see Fig.~\ref{unimodalNet} and Eq~\ref{eqdecreasing}.
\begin{eqnarray}
\label{eqdecreasing}
        z^{r\ell}_K = z_K \nonumber & \\
        z^{r\ell}_i = z^{r\ell}_{i+1} + z_{i}, & \quad i=K-1, \dots, 1
\end{eqnarray}

Setting $z^u_i = \min (z^{\ell r}_i, z^{r\ell}_i)$, then, by construction, $\mathbf{z}^u$ is unimodal.
Finally, setting $\mathbf{\hat y}= \text{softmax}(\mathbf{z}^u)$ provides our unimodal distribution.

\begin{theorem}
Let $(a_n)_{n=1}^K$ be a finite sequence of non-negative real numbers.
Define the sequence $(b_n)$ as 
$$b_n = \sum_{i=1}^n a_i = \begin{cases} a_1 \quad \text{for n=1} \\ b_{n-1}+a_n  \quad  \text{for } n = 2\cdots K\end{cases}$$
Additionally, define the sequence $(c_n)$ as 
$$c_n = \sum_{i=n}^K a_i = \begin{cases} a_K \quad \text{for n=K} \\ c_{n+1}+a_n  \quad  \text{for } n=K-1, \cdots, 1\end{cases}$$
Finally, define $(d_n)$ as $d_n = \min (b_n,c_n)$ and $(y_n) = softmax ((d_n))$. Then, $y_n$ is a unimodal distribution.
 
\end{theorem}

\begin{proof}
Clearly $(b_n)$ is a non-decreasing sequence ($b_n\geq b_{n-1}$), while $(c_n)$ is a non-increasing sequence ($c_n\leq c_{n-1}$).
Let $\ell \in \{1,\dots, K\}$ be the smallest index for which $c_\ell \geq b_\ell$. Then $d_i = b_i$ for $i<\ell$, $d_i = c_i$, otherwise, and clearly unimodal. If $\ell$ does not exist, then $d_i = b_i \ \forall i \in \{1,\dots, K\}$ and also unimodal.

\end{proof}

\subsection{Wasserstein Regularization: a Non-parametric unimodal loss}
\label{sec:soft-proposal}

A regularization term $\text{u}(\delta,{\mathbf y}_n,\mathbf{\hat y}_n)$ has been heuristically introduced in the literature to penalize, non-parametrically, deviations from the family of unimodal distributions. Let $k_n^\star$ be the true label.
Defining $\text{ReLU}(x) = \max(0,x)$, a possible fix for an order-aware loss has been previously proposed~\cite{ordinal_albuquerque2021} as
\begin{equation}
\text{CO2}(\delta,{\mathbf y}_n,\mathbf{\hat y}_n) = \text{CE}({\mathbf y}_n,\mathbf{\hat y}_n) + \lambda\,\text{u}(\delta,{\mathbf y}_n,\mathbf{\hat y}_n).
\end{equation}
where $\lambda \geq 0$ controls the relative influence of the extra term $u$ which favours unimodal distributions and is defined as
\begin{eqnarray}
\text{u}(\delta,{\mathbf y}_n,\mathbf{\hat y}_n) = \sum_{k=1}^{k_n^\star} \text{ReLU}(\delta + \hat y_{nk} - \hat y_{n(k+1)}) +\nonumber\\
\sum_{k=k_n^\star}^{K-1} \text{ReLU}(\delta + \hat y_{n(k+1)} - \hat y_{nk}).
\end{eqnarray}
Furthermore, a margin of $\delta\geq 0$ ensures that the difference between consecutive probabilities is at least $\delta$~\cite{ordinal_albuquerque2021}. As a special case, CO has been defined as the case when the margin is zero ($\delta=0$),
\begin{equation}
\text{CO}({\mathbf y}_n,\mathbf{\hat y}_n)=\text{CO2}(0,{\mathbf y}_n,\mathbf{\hat y}_n).
\end{equation}

Other heuristics are possible. For instance, instead of penalizing the ``errors''  between consecutive values, one could penalize the errors between any pair of probability values for which an order relation is defined, Eq.~\eqref{pairs}.

\begin{eqnarray}
\label{pairs}
\text{uu}(\delta,{\mathbf y}_n,\mathbf{\hat y}_n) = \sum_{\ell=1}^{k_n^\star-1} \sum_{k=\ell+1}^{k_n^\star} \text{ReLU}(\delta + \hat y_{n\ell} - \hat y_{nk}) + \nonumber \\
\sum_{\ell=k_n^\star}^{K-1} \sum_{k=\ell+1}^{K} \text{ReLU}(\delta + \hat y_{nk} - \hat y_{n\ell}).
\end{eqnarray}

As an illustrative example, if $\delta=0$, $\mathbf{\hat y} = \left [2/6 \  3/6  \ 0  \  1/6\right ]^t$, and $k^\star = 4$ then $\text{u}(\delta,{\mathbf y}_n,\mathbf{\hat y}_n) = 3/6$ and  $\text{uu}(\delta,{\mathbf y}_n,\mathbf{\hat y}_n) = 8/6$.
For $\mathbf{\hat y} = [2/6 \  3/6  \ 1/6  \  0]^t$, and $k^\star = 4$ then $\text{u}(\delta,{\mathbf y}_n,\mathbf{\hat y}_n) = 3/6$ and $ \text{uu}(\delta,{\mathbf y}_n,\mathbf{\hat y}_n) = 9/6$.

Departing from these heuristics, we now follow a set-theory approach, supported by the notion of projection in a set. 

Let $\cal S$ be the set of all unimodal distributions with mode in $k^\star$.
Let also ${\cal D}_1(.,.)$ and ${\cal D}_2(.,.)$ be suitable measures of dissimilarity between two distributions (for instance, Kullback-Leibler (KL) divergence or Wasserstein Distance).
The projection in $\cal S$ of a distribution $\mathbf{\hat y}_n$ is defined as

\begin{equation}
\label{eqDistUni}
\mathbf{\hat y}_n^P = \arg\min_{\mathbf y\in {\cal S}} {\cal D}_1({\mathbf y}, \mathbf{\hat y}_n).
\end{equation}
$\mathbf{\hat y}_n^P$ is the unimodal distribution more similar to $\mathbf{\hat y}_n$. 
The loss term during the learning stage is then defined as

\begin{equation}
\label{UniLossTerm}
{\cal D}_2(\mathbf{\hat y}_n^P, \mathbf{\hat y}_n).
\end{equation}

When ${\cal D}_1 = {\cal D}_2 = {\cal D}$, the loss term simplifies to ${\cal D}(\mathbf{\hat y}_n^P, \mathbf{\hat y}_n) = \min_{\mathbf y\in {\cal S}} {\cal D}({\mathbf y}, \mathbf{\hat y}_n)$.
Nevertheless, in practice, for analytical and numerical reasons, we found it useful to also consider  ${\cal D}_1 \not = {\cal D}_2$.
We next analyze the computation of the projection when ${\cal D}_1$ is mapped to the Wasserstein Distance.

Wasserstein distances are metrics on probability distributions inspired by the optimal mass transport problem. Roughly speaking, they measure the minimum effort required to reconfigure the probability mass of one distribution in order to recover the other distribution. They are ubiquitous in mathematics and have a long history as a catalyst for significant developments in analysis, optimization, and probability theory~\cite{Kantorovich1960}.

\medskip\noindent
\textbf{Wasserstein Projection:}
The projection of distribution $\mathbf{\hat y}$ in $\cal S$ (Eq.~\eqref{eqDistUni}) is easily computed by modifying the computation of the Wasserstein Distance between two distributions.

Giving the transportation cost $c_{ij}\geq 0$ between positions $i$ and $j$, the Wasserstein Distance between the two distributions $\mathbf{p}$ and $\mathbf{q}$ is computed as
\begin{eqnarray}
\label{eq:Wasserstein}
  \min_{t_{ij}} \sum_{ij} t_{ij}c_{ij} & \nonumber \\
  s.t. &  \nonumber \\
& \sum_j t_{ij} = p_j, \forall j \in \{1,\dots, K\} \nonumber \\
& \sum_i t_{ij} = q_i, \forall i \in \{1,\dots, K\} \nonumber \\
& t_{ij}\geq 0, \forall i,j
\end{eqnarray}

The distance of $\mathbf{q}$ to the set of unimodal distributions is trivially computed as
\begin{eqnarray}
\label{eq:Wassersteinunimodal}
  \min_{t_{ij}} \sum_{ij} t_{ij}c_{ij} & \nonumber \\
  s.t. & \nonumber \\
& \sum_i t_{ij} = q_j, \forall j \in \{1,\dots, K\}\nonumber \\
& \sum_j t_{ij} \leq \sum_{j+1} t_{ij}, \forall i \in \{1,\dots, k^\star-1\}\nonumber \\
& \sum_j t_{ij} \geq \sum_{j+1} t_{ij}, \forall i \in \{k^\star,\dots, K-1\}\nonumber \\
& t_{ij}\geq 0, \forall i,j
\end{eqnarray}

The projection of $\mathbf{q}$ in ${\cal S}$ results as $\mathbf{q}^P_i=\sum_j t_{ij}$. Equations~\ref{eq:Wasserstein} and~\ref{eq:Wassersteinunimodal} are easily solved with linear optimization, with efficient solutions available for one-dimensional distributions.


\medskip\noindent
\textbf{Training Loss:} The penalty given by Eq.~\eqref{UniLossTerm} cannot be used standalone to train a (deep neural network) model.
The learning would most likely converge to the ``constant model'', always outputting a uniform distribution.
The penalty~\eqref{UniLossTerm} penalizes non-uniform distributions but gives zero penalties to any unimodal distribution.

For that reason, a CE term is added. The weighted sum of the CE with the unimodal penalty simultaneously gives preference to unimodal distributions and, in the set of unimodal distributions, gives preference to the zero-entropy distribution, with ${\hat{y}}_{k^\star}=1$,
\begin{equation}
\mathcal L(\mathbf{y}_n,\mathbf{\hat y}_n)=\underbrace{\text{CE}(\mathbf{y}_n,\mathbf{\hat y}_n)}_\text{main term}+\lambda\underbrace{\mathcal D_2(\mathbf{\hat y}^P_n,\mathbf{\hat y}_n).}_\text{regularization term}
\end{equation}


\section{Experimental Details}
The proposed models are contrasted against previously mentioned methods from the literature across ten datasets, both tabular and image datasets.

\subsection{Methods}
Ten baseline methods are considered, divided in three families: (Non-unimodal methods) Cross-Entropy (CE), Ordinal Encoding (OE)~\cite{cheng2008neural}, CDW-CE~\cite{polat2022class}; (Unimodal hard constraints methods) Binomial Unimodal (BU)~\cite{Costa2005Classification}, Poisson Unimodal (PU)~\cite{Beckham2017} with $\tau=1$, ORD-ACL and VS-SL~\cite{yamasaki2022unimodal}; and (Unimodal soft constraints methods) Unimodal Regularization (UR)~\cite{liu2020unimodal} and a non-parametric unimodal regularizer CO2~\cite{ordinal_albuquerque2021}.

Three proposals are considered: the hard UnimodalNet architecture from section~\ref{sec:hard-proposal} (UN), and two proposed soft regularizations from section~\ref{sec:soft-proposal} (WU-KLDiv and WU-Wass). The difference between the two regularizers is how regularization is performed relative to the projection using the Wasserstein distance (${\cal D}_1$): WU-KLDiv uses Kullback–Leibler divergence for ${\cal D}_2$, while WU-Wass uses the Wasserstein distance for ${\cal D}_2$.

\subsection{Datasets used for the Experiments}

\begin{table}
\setlength{\tabcolsep}{2.5pt}  
\caption{Datasets}
\label{table:datasets}
\begin{tabular}{lp{8em}rrr}
\hline
\multicolumn{4}{c}{\bf Tabular datasets} \\
\hline
\bf Name & \bf Description & \bf N & \bf K & \bf IR \\
Abalone5~\cite{UCI} & Predict the age of abalone from physical measurements (discretized in 5 classes) & 4,177 & 5 & 32.1 \\
Abalone10~\cite{UCI} & \textit{idem} (discretized in 10 classes) & 4,177 & 10 & 32.3 \\
Balance scale~\cite{UCI} & Balance scale weight \& distance database & 625 & 3 & 1.0 \\
Car~\cite{UCI} & Car qualitative evaluation & 1,728 & 4 & 4.5 \\
New thyroid~\cite{UCI} & Normal/hyper/hypo-thyroidism & 215 & 3 & 2.0 \\
\hline
\multicolumn{3}{c}{\bf Image datasets} \\
\hline
\bf Name & \bf Description & \bf N & \bf K & \bf IR \\
BACH~\cite{BACH} & Breast histology microscopy & 400 & 4 & 1.0 \\
DHCI~\cite{HCI} & Dating historical color images & 1,325 & 5 & 1.0 \\
FGNET~\cite{FGNET} & Age estimation & 1,002 & 70 & 3.7 \\
Pap Smear~\cite{PAPSMEAR} & Pap Smear from DTU/Herlev & 570 & 5 & 1.2 \\
FocusPath~\cite{FocusPath} & Estimate focus level for whole-slide images & 8,640 & 12 & 3.2 \\
\hline
\end{tabular}
\end{table}

Five tabular and five image datasets were used for evaluation, as detailed in Table~\ref{table:datasets}. The tabular datasets come from the UCI Machine Learning Repository~\cite{UCI}: Abalone (discretized into 5 and 10 classes), Balance Scale, Car Evaluation, and New Thyroid; while the image datasets consist of BACH (breast histology microscopy), DHCI (dating historical images), FGNET (age estimation), Pap Smear, and FocusPath (focus level estimation in whole-slide images). These datasets were chosen because they cover a range of real-world ordinal regression tasks, such as age estimation, disease classification, and image focus prediction, which benefit from considering the inherent ordering in the data.

For evaluation purposes, a stratified 5-fold split was performed: the first fold was used for hyperparameter validation, while the remaining four folds were used to estimate the final results.

For the tabular datasets, the exogenous variables were z-normalized ($x'=\frac{x-\mu}{\sigma}$), and categorical variables were one-hot encoded. For the image datasets, the images were normalized using the same normalization as ImageNet, and the images were resized to $268\times268$; the pre-processing was followed by data augmentation: random crop to $256\times256$, brightness and contrast jitter of $[-10\%,10\%]$, and horizontal flipping (with $p=0.5$).

The imbalance ratio (IR) column~\cite{perez2014graph} illustrates how imbalance the individual data sets are, where 1=balanced and the higher the value, the greater the imbalance,
\begin{equation}
\text{IR}=\frac{1}{K}\sum_{k=1}^K\frac{N-N_k}{(K-1)N_k}.
\end{equation}

\subsection{Architectures}

The following neural network architectures were used. For the tabular data, an MLP with 128 hidden neurons and a ReLU activation function was trained for 1,000 epochs. While for images, a ResNet-18~\cite{resnet} pre-trained in ImageNet, trained for 100 epochs. Adam was used as the optimizer with a learning rate of $10^{-4}$.

For each loss with a regularization term (WU-KLDiv, WU-Wass and CO2), the regularization coefficient ($\lambda$) was validated across the linear space $\lambda\in\{10^{-3},10^{-2},\dots,10^3\}$. For all other hyperparameters, the default values suggested by the authors were used.

\subsection{Evaluation Metrics}
Four classical metrics for ordinality are used: (i)~Accuracy, (ii)~Mean Absolute Error (MAE), (iii)~Quadratic Weighted Kappa (QWK), and (iv)~Kendall's $\tau$. While accuracy is categorical and MAE is a regression metric, they are both often used in tandem for ordinal problems since an ordinal problem is a mix of both problems. QWK and Kendall's $\tau$ are other popular metrics for ordinal regression as they measure agreement -- agreement weighted by the square class distance in the case of QWK and agreement between the rank correlations in the case of Kendall's $\tau$.

Furthermore, a ``\%Unimodality'' column shows the fraction of times that the probability distribution produced by the model is unimodal. And in order to detect biases, Zero Mean Error (ZME) is shown, which is the residual average, and the Negative Log-Likelihood (NLL) score.

\section{Results and Discussion}

\begin{table}
\centering
\caption{Average Summary of the Results (with \textbf{bold} for the best in each group).}
\footnotesize
\label{table:results-average}

\begin{tabular}{l|rrrrrrr}
\hline
\multicolumn{8}{c}{\bf All Datasets (10)} \\
\bf Method & Acc & QWK & $\tau$ & MAE & \%Uni & ZME & NLL \\
\hline
\multicolumn{8}{l}{Non-unimodal methods} \\
CE & 66.1 & 74.8 & 71.8 & 0.94 & 76.3 & -0.32 & 1.19 \\
POM & 54.4 & 71.7 & 68.2 & 1.80 & 48.0 & -1.21 & 2.00 \\
OE & \bf68.0 & \bf80.0 & \bf76.2 & \bf0.63 & 78.3 & \bf-0.06 & \bf0.94 \\
CDW & 57.0 & 74.7 & 72.1 & 0.99 & \bf91.0 & 0.13 & 2.58 \\
\hline
\multicolumn{8}{l}{Unimodal hard constraints methods} \\
BU & 62.2 & \bf79.5 & 73.9 & 0.72 & \bf100 & \bf-0.05 & 1.16 \\
PU & 46.3 & 72.5 & 69.5 & 0.90 & \bf100 & -0.20 & 1.22 \\
ORD-ACL & 64.5 & 75.0 & 72.5 & 0.73 & \bf100 & -0.06 & 1.15 \\
VS-SL & 63.8 & 76.7 & 73.0 & 0.74 & \bf100 & -0.08 & 1.17 \\
UN* & \bf67.0 & 78.1 & \bf74.3 & \bf0.67 & \bf100 & -0.06 & \bf1.01 \\
\hline
\multicolumn{8}{l}{Unimodal soft constraints methods} \\
UR & \bf67.5 & \bf78.0 & 74.4 & 0.73 & 88.1 & \bf-0.16 & 1.08 \\
CO2 & 66.5 & 77.9 & \bf74.7 & \bf0.71 & \bf89.9 & -0.18 & 1.17 \\
WU-KLDIV* & 66.4 & 77.4 & 73.6 & 0.75 & 81.0 & -0.17 & 1.00 \\
WU-Wass* & 66.7 & 77.7 & 74.1 & \bf0.71 & 83.5 & -0.22 & \bf0.98 \\
\hline
\multicolumn{8}{c}{\bf IR\textgreater1.5 Datasets (6)} \\
\bf Method & Acc & QWK & $\tau$ & MAE & \%Uni & ZME & NLL \\
\hline
\multicolumn{8}{l}{Non-unimodal methods} \\
CE & 69.9 & 77.3 & 76.8 & 1.17 & 77.9 & -0.47 & 1.12 \\
POM & 62.5 & 73.2 & 72.0 & 2.48 & 62.4 & -2.08 & 1.93 \\
OE & \bf71.2 & \bf82.9 & \bf81.2 & \bf0.72 & 79.5 & \bf-0.07 & \bf0.86 \\
CDW & 58.7 & 77.2 & 75.6 & 1.28 & \bf85.1 & 0.22 & 2.40 \\
\hline
\multicolumn{8}{l}{Unimodal hard constraints methods} \\
BU & 64.2 & 81.3 & 77.3 & 0.84 & \bf100 & -0.08 & 1.22 \\
PU & 45.8 & 73.4 & 73.0 & 1.02 & \bf100 & -0.36 & 1.30 \\
ORD-ACL & 67.6 & 79.5 & 80.3 & 0.80 & \bf100 & -0.13 & 1.04 \\
VS-SL & 69.5 & 81.1 & 79.9 & 0.81 & \bf100 & -0.15 & 1.00 \\
UN* & \bf70.6 & \bf82.1 & \bf80.5 & \bf0.74 & \bf100 & \bf-0.07 & \bf0.92 \\
\hline
\multicolumn{8}{l}{Unimodal soft constraints methods} \\
UR & \bf70.9 & 81.1 & 79.5 & 0.85 & 86.4 & \bf-0.22 & 1.11 \\
CO2 & 69.7 & 80.8 & \bf79.9 & 0.83 & \bf90.2 & -0.25 & 1.19 \\
WU-KLDIV* & 70.2 & 80.9 & 79.3 & 0.86 & 80.9 & -0.26 & 0.89 \\
WU-Wass* & 69.6 & \bf81.8 & \bf79.9 & \bf0.81 & 86.9 & -0.32 & \bf0.87 \\
\hline
\end{tabular}

\smallskip\raggedright
* proposed model
\end{table}

\begin{figure}
\centering
\begin{tikzpicture}[font=\footnotesize]
\begin{groupplot}[
    group style={
        group name=myplots,
        columns=2,
        horizontal sep=5pt,
    },
    grid=both,
    height=5cm,
    table/col sep=space,
    scale only axis=true,  
    xticklabel={\pgfmathparse{\tick}\pgfmathprintnumber{\pgfmathresult}\%},
    yticklabel={\pgfmathparse{\tick}\pgfmathprintnumber{\pgfmathresult}\%},
    legend pos=south east,
    legend cell align={left},
    legend style={draw=none, fill=none},
]
\nextgroupplot[
    width=0.33*6cm,
    axis y line=left,
    axis line style={-},
    xmin=0, xmax=25,
    ymin=0, ymax=100,
    xtick={0,20},
    ylabel={\%Unimodal},
]
\addplot[
    only marks,
    scatter,
    scatter/classes={1={mark=o}, 2={mark=halfcircle}, 3={fill=gray}, 4={mark=otimes}},
    scatter src=explicit,
    visualization depends on=\thisrow{lambda}\as\lambda,
    scatter/@pre marker code/.append style={/tikz/mark size=(log10(\lambda)+4)/2},
] table[x=acc, y=unimodality, meta=dataset] {data.csv};

\nextgroupplot[
    width=0.66*6cm,
    axis y line=right,
    axis line style={-},
    xmin=55, xmax=100,
    ymin=0, ymax=100,
    yticklabels={},
    xtick={60,80,100},
]
\addplot[
    only marks,
    scatter,
    scatter/classes={1={mark=o}, 2={mark=halfcircle}, 3={fill=gray}, 4={mark=otimes}, 5={mark=square}},
    scatter src=explicit,
    visualization depends on=\thisrow{lambda}\as\lambda,
    scatter/@pre marker code/.append style={/tikz/mark size=(log10(\lambda)+4)/2},
] table[x=acc, y=unimodality, meta=dataset] {data.csv};

\draw[black, fill=white] (56, 2) rectangle (100, 48);
\legend{ICIAR, DHCI, FGNET, SMEAR2005, FOCUSPATH}

\end{groupplot}
\node[fit=(myplots c1r1)(myplots c2r1)](mygroup){};
\node[below=4mm of mygroup.south] {Accuracy};

\coordinate (lambda1) at ([xshift=-12em, yshift=21.5mm]myplots c2r1.south east);
\coordinate (lambda2) at ([xshift=-12em, yshift=18.5mm]myplots c2r1.south east);
\coordinate (lambda3) at ([xshift=-12em, yshift=15.5mm]myplots c2r1.south east);
\coordinate (lambda4) at ([xshift=-12em, yshift=12.5mm]myplots c2r1.south east);
\coordinate (lambda5) at ([xshift=-12em, yshift=9.5mm]myplots c2r1.south east);
\coordinate (lambda6) at ([xshift=-12em, yshift=6.5mm]myplots c2r1.south east);
\coordinate (lambda7) at ([xshift=-12em, yshift=3.5mm]myplots c2r1.south east);

\draw (lambda1) circle (4.2pt); \node[right=2mm of lambda1, font=\footnotesize] {$\lambda=10^3$};
\draw (lambda2) circle (3.5pt); \node[right=2mm of lambda2, font=\footnotesize] {$\lambda=10^2$};
\draw (lambda3) circle (3pt); \node[right=2mm of lambda3, font=\footnotesize] {$\lambda=10$};
\draw (lambda4) circle (2.1pt); \node[right=2mm of lambda4, font=\footnotesize] {$\lambda=1$};
\draw (lambda5) circle (1.5pt); \node[right=2mm of lambda5, font=\footnotesize] {$\lambda=10^{-1}$};
\draw (lambda6) circle (1pt); \node[right=2mm of lambda6, font=\footnotesize] {$\lambda=10^{-2}$};
\draw (lambda7) circle (0.6pt); \node[right=2mm of lambda7, font=\footnotesize] {$\lambda=10^{-3}$};

\draw ([xshift=-1mm, yshift=-2mm]myplots c1r1.south east) -- ([xshift=1mm, yshift=2mm]myplots c1r1.south east);
\draw ([xshift=-1mm, yshift=-2mm]myplots c2r1.south west) -- ([xshift=1mm, yshift=2mm]myplots c2r1.south west);
\draw ([xshift=-1mm, yshift=-2mm]myplots c1r1.north east) -- ([xshift=1mm, yshift=2mm]myplots c1r1.north east);
\draw ([xshift=-1mm, yshift=-2mm]myplots c2r1.north west) -- ([xshift=1mm, yshift=2mm]myplots c2r1.north west);

\end{tikzpicture}
\caption{Results for different values of $\lambda$ for the proposed WU-Wass. As $\lambda$ increases, output probabilities tend to be more unimodal while sacrificing a little accuracy.}
\label{figure:lambdas}
\end{figure}

\begin{table*}[p]
\vspace{-20ex}
\centering
\footnotesize
\setlength{\tabcolsep}{1pt}  
\caption{Results for Tabular Datasets (with the best method for each metric/dataset in \textbf{bold}).}
\label{table:results-tabular}
\rotatebox{90}{
\begin{tabular}{l|llll|lllll|llll}
\hline\multicolumn{1}{l|}{\bf Dataset} & \multicolumn{4}{c|}{\bf Non-unimodal methods} & \multicolumn{5}{c|}{\bf Unimodal hard constraints methods} & \multicolumn{4}{c}{\bf Unimodal soft constraints methods} \\\hline
\bf Abalone5 & CE & POM & OE & CDW & BU & PU & ORD-ACL & VS-SL & UN* & UR & CO2 & WU-KLDIV* & WU-Wass* \\
\%Accuracy & $78.5\color{gray}\pm0.8$ & $78.6\color{gray}\pm0.8$ & $\mathbf{79.0}\color{gray}\pm1.0$ & $73.9\color{gray}\pm0.9$ & $75.6\color{gray}\pm1.3$ & $28.0\color{gray}\pm2.5$ & $61.3\color{gray}\pm33.7$ & $78.6\color{gray}\pm1.1$ & $78.4\color{gray}\pm0.8$ & $78.6\color{gray}\pm0.9$ & $78.6\color{gray}\pm1.5$ & $\mathbf{79.0}\color{gray}\pm0.8$ & $78.7\color{gray}\pm0.5$ \\
MAE & $0.23\color{gray}\pm0.01$ & $0.23\color{gray}\pm0.01$ & $\mathbf{0.22}\color{gray}\pm0.01$ & $0.27\color{gray}\pm0.01$ & $0.26\color{gray}\pm0.01$ & $0.78\color{gray}\pm0.03$ & $0.45\color{gray}\pm0.43$ & $0.23\color{gray}\pm0.01$ & $0.23\color{gray}\pm0.01$ & $0.23\color{gray}\pm0.01$ & $0.23\color{gray}\pm0.01$ & $\mathbf{0.22}\color{gray}\pm0.01$ & $0.23\color{gray}\pm0.00$ \\
QWK & $53.9\color{gray}\pm1.7$ & $56.6\color{gray}\pm2.1$ & $57.3\color{gray}\pm2.0$ & $58.1\color{gray}\pm2.8$ & $\mathbf{58.5}\color{gray}\pm2.3$ & $32.1\color{gray}\pm2.2$ & $41.6\color{gray}\pm27.8$ & $54.1\color{gray}\pm2.2$ & $55.3\color{gray}\pm1.8$ & $54.2\color{gray}\pm0.9$ & $54.4\color{gray}\pm2.7$ & $55.5\color{gray}\pm1.6$ & $54.8\color{gray}\pm0.7$ \\
\%$\tau$ & $55.9\color{gray}\pm1.5$ & $57.5\color{gray}\pm2.0$ & $58.7\color{gray}\pm1.9$ & $57.4\color{gray}\pm2.4$ & $59.3\color{gray}\pm1.9$ & $48.2\color{gray}\pm1.1$ & $57.0\color{gray}\pm1.4$ & $56.4\color{gray}\pm2.2$ & $57.0\color{gray}\pm1.2$ & $56.2\color{gray}\pm1.3$ & $56.3\color{gray}\pm3.4$ & $57.1\color{gray}\pm1.7$ & $56.6\color{gray}\pm0.7$ \\
\%Unimodal & $99.8\color{gray}\pm0.1$ & $\mathbf{100}\color{gray}\pm0.0$ & $99.7\color{gray}\pm0.1$ & $\mathbf{100}\color{gray}\pm0.0$ & $\mathbf{100}\color{gray}\pm0.0$ & $\mathbf{100}\color{gray}\pm0.0$ & $\mathbf{100}\color{gray}\pm0.0$ & $\mathbf{100}\color{gray}\pm0.0$ & $\mathbf{100}\color{gray}\pm0.0$ & $\mathbf{100}\color{gray}\pm0.1$ & $\mathbf{100}\color{gray}\pm0.1$ & $99.8\color{gray}\pm0.2$ & $99.7\color{gray}\pm0.2$ \\
ZME & $-0.07\color{gray}\pm0.02$ & $-0.07\color{gray}\pm0.01$ & $\mathbf{-0.06}\color{gray}\pm0.01$ & $0.07\color{gray}\pm0.01$ & $-0.15\color{gray}\pm0.01$ & $-0.73\color{gray}\pm0.02$ & $-0.32\color{gray}\pm0.52$ & $\mathbf{-0.06}\color{gray}\pm0.00$ & $\mathbf{-0.06}\color{gray}\pm0.02$ & $-0.09\color{gray}\pm0.01$ & $\mathbf{-0.06}\color{gray}\pm0.02$ & $-0.07\color{gray}\pm0.00$ & $-0.07\color{gray}\pm0.01$ \\
NLL & $0.53\color{gray}\pm0.02$ & $0.53\color{gray}\pm0.02$ & $\mathbf{0.52}\color{gray}\pm0.02$ & $0.55\color{gray}\pm0.02$ & $0.94\color{gray}\pm0.01$ & $1.14\color{gray}\pm0.01$ & $0.54\color{gray}\pm0.04$ & $0.56\color{gray}\pm0.01$ & $0.53\color{gray}\pm0.01$ & $0.81\color{gray}\pm0.00$ & $1.15\color{gray}\pm0.01$ & $0.53\color{gray}\pm0.02$ & $0.53\color{gray}\pm0.02$ \\
\hline
\bf Abalone10 & CE & POM & OE & CDW & BU & PU & ORD-ACL & VS-SL & UN* & UR & CO2 & WU-KLDIV* & WU-Wass* \\
\%Accuracy & $\mathbf{58.0}\color{gray}\pm2.3$ & $57.5\color{gray}\pm2.0$ & $57.6\color{gray}\pm2.2$ & $44.2\color{gray}\pm1.7$ & $53.0\color{gray}\pm3.1$ & $38.2\color{gray}\pm1.5$ & $57.4\color{gray}\pm2.1$ & $56.3\color{gray}\pm1.9$ & $57.3\color{gray}\pm2.5$ & $\mathbf{58.0}\color{gray}\pm2.7$ & $55.6\color{gray}\pm1.8$ & $57.9\color{gray}\pm2.6$ & $57.9\color{gray}\pm2.4$ \\
MAE & $0.53\color{gray}\pm0.02$ & $\mathbf{0.52}\color{gray}\pm0.03$ & $\mathbf{0.52}\color{gray}\pm0.02$ & $0.66\color{gray}\pm0.02$ & $0.56\color{gray}\pm0.04$ & $0.77\color{gray}\pm0.01$ & $0.54\color{gray}\pm0.02$ & $0.56\color{gray}\pm0.03$ & $0.54\color{gray}\pm0.03$ & $0.53\color{gray}\pm0.03$ & $0.53\color{gray}\pm0.02$ & $0.54\color{gray}\pm0.02$ & $0.53\color{gray}\pm0.02$ \\
QWK & $62.3\color{gray}\pm2.0$ & $66.0\color{gray}\pm2.6$ & $66.6\color{gray}\pm1.4$ & $57.9\color{gray}\pm1.8$ & $\mathbf{69.1}\color{gray}\pm1.8$ & $60.6\color{gray}\pm2.4$ & $63.9\color{gray}\pm2.3$ & $60.6\color{gray}\pm2.3$ & $64.8\color{gray}\pm3.2$ & $62.3\color{gray}\pm1.4$ & $63.9\color{gray}\pm1.4$ & $62.4\color{gray}\pm2.1$ & $63.3\color{gray}\pm1.6$ \\
\%$\tau$ & $63.1\color{gray}\pm2.5$ & $64.5\color{gray}\pm1.7$ & $64.6\color{gray}\pm2.2$ & $60.9\color{gray}\pm2.0$ & $64.9\color{gray}\pm2.2$ & $64.6\color{gray}\pm1.5$ & $63.4\color{gray}\pm2.3$ & $62.4\color{gray}\pm1.8$ & $63.4\color{gray}\pm2.2$ & $63.3\color{gray}\pm1.9$ & $64.1\color{gray}\pm1.6$ & $63.1\color{gray}\pm2.5$ & $63.2\color{gray}\pm2.1$ \\
\%Unimodal & $91.1\color{gray}\pm1.0$ & $96.6\color{gray}\pm0.8$ & $92.4\color{gray}\pm1.8$ & $99.9\color{gray}\pm0.1$ & $\mathbf{100}\color{gray}\pm0.0$ & $\mathbf{100}\color{gray}\pm0.0$ & $\mathbf{100}\color{gray}\pm0.0$ & $\mathbf{100}\color{gray}\pm0.0$ & $\mathbf{100}\color{gray}\pm0.0$ & $98.5\color{gray}\pm0.4$ & $\mathbf{100}\color{gray}\pm0.0$ & $91.3\color{gray}\pm0.6$ & $91.5\color{gray}\pm2.1$ \\
ZME & $-0.18\color{gray}\pm0.02$ & $-0.10\color{gray}\pm0.03$ & $-0.10\color{gray}\pm0.02$ & $0.25\color{gray}\pm0.04$ & $-0.19\color{gray}\pm0.02$ & $-0.61\color{gray}\pm0.02$ & $-0.14\color{gray}\pm0.05$ & $-0.15\color{gray}\pm0.04$ & $-0.13\color{gray}\pm0.07$ & $-0.19\color{gray}\pm0.02$ & $\mathbf{0.01}\color{gray}\pm0.01$ & $-0.18\color{gray}\pm0.02$ & $-0.17\color{gray}\pm0.02$ \\
NLL & $1.02\color{gray}\pm0.03$ & $1.04\color{gray}\pm0.03$ & $\mathbf{1.01}\color{gray}\pm0.02$ & $1.54\color{gray}\pm0.15$ & $1.39\color{gray}\pm0.01$ & $1.60\color{gray}\pm0.01$ & $1.03\color{gray}\pm0.05$ & $1.03\color{gray}\pm0.04$ & $1.03\color{gray}\pm0.03$ & $1.21\color{gray}\pm0.01$ & $1.51\color{gray}\pm0.01$ & $\mathbf{1.01}\color{gray}\pm0.02$ & $1.02\color{gray}\pm0.02$ \\
\hline
\bf Balance scale & CE & POM & OE & CDW & BU & PU & ORD-ACL & VS-SL & UN* & UR & CO2 & WU-KLDIV* & WU-Wass* \\
\%Accuracy & $23.8\color{gray}\pm3.2$ & $32.2\color{gray}\pm1.6$ & $32.4\color{gray}\pm6.6$ & $28.4\color{gray}\pm3.3$ & $\mathbf{33.4}\color{gray}\pm3.7$ & $31.4\color{gray}\pm2.1$ & $31.0\color{gray}\pm4.8$ & $32.6\color{gray}\pm3.3$ & $29.4\color{gray}\pm4.7$ & $25.0\color{gray}\pm2.6$ & $30.6\color{gray}\pm3.8$ & $25.6\color{gray}\pm4.3$ & $25.0\color{gray}\pm4.1$ \\
MAE & $1.22\color{gray}\pm0.09$ & $0.96\color{gray}\pm0.05$ & $\mathbf{0.93}\color{gray}\pm0.09$ & $0.94\color{gray}\pm0.05$ & $\mathbf{0.93}\color{gray}\pm0.08$ & $1.08\color{gray}\pm0.08$ & $1.15\color{gray}\pm0.08$ & $1.13\color{gray}\pm0.08$ & $1.11\color{gray}\pm0.07$ & $1.16\color{gray}\pm0.05$ & $\mathbf{0.93}\color{gray}\pm0.06$ & $1.15\color{gray}\pm0.07$ & $1.14\color{gray}\pm0.10$ \\
QWK & $44.5\color{gray}\pm7.3$ & $\mathbf{55.3}\color{gray}\pm4.8$ & $52.4\color{gray}\pm6.4$ & $44.4\color{gray}\pm5.3$ & $\mathbf{55.3}\color{gray}\pm5.9$ & $52.3\color{gray}\pm6.9$ & $39.7\color{gray}\pm6.8$ & $40.2\color{gray}\pm5.3$ & $46.1\color{gray}\pm4.8$ & $46.6\color{gray}\pm6.1$ & $52.4\color{gray}\pm4.4$ & $46.2\color{gray}\pm4.9$ & $43.7\color{gray}\pm9.4$ \\
\%$\tau$ & $37.2\color{gray}\pm5.7$ & $45.3\color{gray}\pm4.7$ & $44.4\color{gray}\pm5.6$ & $46.5\color{gray}\pm5.4$ & $46.1\color{gray}\pm5.4$ & $42.7\color{gray}\pm5.8$ & $31.8\color{gray}\pm6.0$ & $33.6\color{gray}\pm4.6$ & $37.9\color{gray}\pm4.3$ & $38.7\color{gray}\pm4.3$ & $46.2\color{gray}\pm4.4$ & $37.5\color{gray}\pm3.9$ & $35.9\color{gray}\pm7.3$ \\
\%Unimodal & $74.8\color{gray}\pm3.5$ & $43.6\color{gray}\pm5.6$ & $73.6\color{gray}\pm4.9$ & $\mathbf{100}\color{gray}\pm0.0$ & $\mathbf{100}\color{gray}\pm0.0$ & $\mathbf{100}\color{gray}\pm0.0$ & $\mathbf{100}\color{gray}\pm0.0$ & $\mathbf{100}\color{gray}\pm0.0$ & $\mathbf{100}\color{gray}\pm0.0$ & $90.6\color{gray}\pm4.5$ & $\mathbf{100}\color{gray}\pm0.0$ & $90.4\color{gray}\pm4.2$ & $82.4\color{gray}\pm2.8$ \\
ZME & $-0.08\color{gray}\pm0.17$ & $\mathbf{-0.00}\color{gray}\pm0.10$ & $-0.07\color{gray}\pm0.08$ & $0.02\color{gray}\pm0.06$ & $-0.01\color{gray}\pm0.08$ & $0.11\color{gray}\pm0.09$ & $0.05\color{gray}\pm0.10$ & $0.19\color{gray}\pm0.13$ & $0.05\color{gray}\pm0.19$ & $-0.02\color{gray}\pm0.22$ & $-0.04\color{gray}\pm0.10$ & $\mathbf{-0.00}\color{gray}\pm0.07$ & $-0.07\color{gray}\pm0.11$ \\
NLL & $1.49\color{gray}\pm0.07$ & $1.45\color{gray}\pm0.04$ & $1.47\color{gray}\pm0.08$ & $6.49\color{gray}\pm0.08$ & $1.48\color{gray}\pm0.07$ & $\mathbf{1.39}\color{gray}\pm0.08$ & $1.40\color{gray}\pm0.06$ & $1.42\color{gray}\pm0.12$ & $\mathbf{1.39}\color{gray}\pm0.08$ & $1.44\color{gray}\pm0.04$ & $1.50\color{gray}\pm0.02$ & $1.42\color{gray}\pm0.06$ & $1.45\color{gray}\pm0.06$ \\
\hline
\bf Car & CE & POM & OE & CDW & BU & PU & ORD-ACL & VS-SL & UN* & UR & CO2 & WU-KLDIV* & WU-Wass* \\
\%Accuracy & $\mathbf{100}\color{gray}\pm0.0$ & $99.3\color{gray}\pm0.5$ & $99.9\color{gray}\pm0.1$ & $95.8\color{gray}\pm0.5$ & $90.4\color{gray}\pm1.1$ & $80.3\color{gray}\pm1.4$ & $99.5\color{gray}\pm0.5$ & $97.4\color{gray}\pm1.2$ & $\mathbf{100}\color{gray}\pm0.0$ & $99.6\color{gray}\pm0.7$ & $\mathbf{100}\color{gray}\pm0.0$ & $99.9\color{gray}\pm0.1$ & $99.9\color{gray}\pm0.1$ \\
MAE & $\mathbf{0.00}\color{gray}\pm0.00$ & $0.01\color{gray}\pm0.01$ & $\mathbf{0.00}\color{gray}\pm0.00$ & $0.04\color{gray}\pm0.00$ & $0.10\color{gray}\pm0.01$ & $0.20\color{gray}\pm0.01$ & $0.01\color{gray}\pm0.01$ & $0.03\color{gray}\pm0.01$ & $\mathbf{0.00}\color{gray}\pm0.00$ & $\mathbf{0.00}\color{gray}\pm0.01$ & $\mathbf{0.00}\color{gray}\pm0.00$ & $\mathbf{0.00}\color{gray}\pm0.00$ & $\mathbf{0.00}\color{gray}\pm0.00$ \\
QWK & $\mathbf{100}\color{gray}\pm0.0$ & $99.3\color{gray}\pm0.5$ & $99.9\color{gray}\pm0.1$ & $95.1\color{gray}\pm0.4$ & $91.2\color{gray}\pm1.7$ & $83.7\color{gray}\pm0.9$ & $99.2\color{gray}\pm0.8$ & $97.7\color{gray}\pm1.0$ & $\mathbf{100}\color{gray}\pm0.0$ & $99.6\color{gray}\pm0.6$ & $\mathbf{100}\color{gray}\pm0.0$ & $99.9\color{gray}\pm0.1$ & $99.9\color{gray}\pm0.1$ \\
\%$\tau$ & $100.0\color{gray}\pm0.0$ & $99.4\color{gray}\pm0.7$ & $99.9\color{gray}\pm0.3$ & $99.0\color{gray}\pm0.3$ & $86.0\color{gray}\pm1.3$ & $70.2\color{gray}\pm4.5$ & $99.7\color{gray}\pm0.3$ & $97.3\color{gray}\pm0.9$ & $100.0\color{gray}\pm0.0$ & $99.7\color{gray}\pm0.4$ & $100.0\color{gray}\pm0.0$ & $99.9\color{gray}\pm0.3$ & $99.9\color{gray}\pm0.3$ \\
\%Unimodal & $99.8\color{gray}\pm0.1$ & $\mathbf{100}\color{gray}\pm0.0$ & $\mathbf{100}\color{gray}\pm0.0$ & $99.9\color{gray}\pm0.1$ & $\mathbf{100}\color{gray}\pm0.0$ & $\mathbf{100}\color{gray}\pm0.0$ & $\mathbf{100}\color{gray}\pm0.0$ & $\mathbf{100}\color{gray}\pm0.0$ & $\mathbf{100}\color{gray}\pm0.0$ & $99.8\color{gray}\pm0.1$ & $99.8\color{gray}\pm0.1$ & $99.8\color{gray}\pm0.1$ & $99.8\color{gray}\pm0.1$ \\
ZME & $\mathbf{0.00}\color{gray}\pm0.00$ & $\mathbf{-0.00}\color{gray}\pm0.01$ & $\mathbf{0.00}\color{gray}\pm0.00$ & $-0.03\color{gray}\pm0.01$ & $-0.03\color{gray}\pm0.01$ & $-0.10\color{gray}\pm0.04$ & $0.01\color{gray}\pm0.00$ & $0.02\color{gray}\pm0.02$ & $\mathbf{0.00}\color{gray}\pm0.00$ & $\mathbf{0.00}\color{gray}\pm0.01$ & $\mathbf{0.00}\color{gray}\pm0.00$ & $\mathbf{0.00}\color{gray}\pm0.00$ & $\mathbf{0.00}\color{gray}\pm0.00$ \\
NLL & $\mathbf{0.00}\color{gray}\pm0.00$ & $0.08\color{gray}\pm0.01$ & $\mathbf{0.00}\color{gray}\pm0.00$ & $0.01\color{gray}\pm0.01$ & $0.25\color{gray}\pm0.01$ & $0.32\color{gray}\pm0.01$ & $0.03\color{gray}\pm0.03$ & $0.14\color{gray}\pm0.07$ & $\mathbf{0.00}\color{gray}\pm0.00$ & $0.43\color{gray}\pm0.00$ & $\mathbf{0.00}\color{gray}\pm0.00$ & $\mathbf{0.00}\color{gray}\pm0.00$ & $\mathbf{0.00}\color{gray}\pm0.00$ \\
\hline
\bf New thyroid & CE & POM & OE & CDW & BU & PU & ORD-ACL & VS-SL & UN* & UR & CO2 & WU-KLDIV* & WU-Wass* \\
\%Accuracy & $94.8\color{gray}\pm4.8$ & $85.5\color{gray}\pm4.0$ & $\mathbf{95.3}\color{gray}\pm5.0$ & $84.9\color{gray}\pm9.8$ & $89.0\color{gray}\pm4.8$ & $83.7\color{gray}\pm5.7$ & $\mathbf{95.3}\color{gray}\pm5.0$ & $90.7\color{gray}\pm6.8$ & $94.8\color{gray}\pm4.8$ & $\mathbf{95.3}\color{gray}\pm5.0$ & $94.8\color{gray}\pm4.8$ & $94.8\color{gray}\pm4.8$ & $\mathbf{95.3}\color{gray}\pm3.8$ \\
MAE & $0.09\color{gray}\pm0.09$ & $0.18\color{gray}\pm0.08$ & $\mathbf{0.08}\color{gray}\pm0.09$ & $0.15\color{gray}\pm0.10$ & $0.14\color{gray}\pm0.07$ & $0.19\color{gray}\pm0.08$ & $\mathbf{0.08}\color{gray}\pm0.09$ & $0.10\color{gray}\pm0.07$ & $0.09\color{gray}\pm0.09$ & $\mathbf{0.08}\color{gray}\pm0.09$ & $0.09\color{gray}\pm0.09$ & $0.09\color{gray}\pm0.09$ & $\mathbf{0.08}\color{gray}\pm0.07$ \\
QWK & $82.7\color{gray}\pm20.5$ & $75.6\color{gray}\pm17.6$ & $83.3\color{gray}\pm20.7$ & $81.4\color{gray}\pm14.0$ & $79.8\color{gray}\pm15.0$ & $77.5\color{gray}\pm13.9$ & $83.3\color{gray}\pm20.7$ & $\mathbf{86.5}\color{gray}\pm8.2$ & $82.7\color{gray}\pm20.5$ & $84.8\color{gray}\pm21.8$ & $82.7\color{gray}\pm20.5$ & $82.7\color{gray}\pm20.5$ & $86.0\color{gray}\pm14.2$ \\
\%$\tau$ & $85.6\color{gray}\pm14.9$ & $74.0\color{gray}\pm10.4$ & $86.7\color{gray}\pm15.3$ & $84.5\color{gray}\pm12.9$ & $79.5\color{gray}\pm8.6$ & $76.8\color{gray}\pm8.4$ & $86.7\color{gray}\pm15.3$ & $88.2\color{gray}\pm7.9$ & $85.6\color{gray}\pm14.9$ & $87.2\color{gray}\pm15.8$ & $85.6\color{gray}\pm14.9$ & $85.6\color{gray}\pm14.9$ & $87.6\color{gray}\pm11.1$ \\
\%Unimodal & $76.2\color{gray}\pm2.2$ & $73.3\color{gray}\pm4.0$ & $75.0\color{gray}\pm8.4$ & $\mathbf{100}\color{gray}\pm0.0$ & $\mathbf{100}\color{gray}\pm0.0$ & $\mathbf{100}\color{gray}\pm0.0$ & $\mathbf{100}\color{gray}\pm0.0$ & $\mathbf{100}\color{gray}\pm0.0$ & $\mathbf{100}\color{gray}\pm0.0$ & $94.2\color{gray}\pm1.3$ & $74.4\color{gray}\pm5.7$ & $77.9\color{gray}\pm4.0$ & $74.4\color{gray}\pm5.7$ \\
ZME & $-0.04\color{gray}\pm0.12$ & $-0.09\color{gray}\pm0.10$ & $-0.05\color{gray}\pm0.12$ & $-0.07\color{gray}\pm0.11$ & $-0.08\color{gray}\pm0.10$ & $-0.04\color{gray}\pm0.08$ & $-0.05\color{gray}\pm0.12$ & $\mathbf{-0.01}\color{gray}\pm0.10$ & $-0.04\color{gray}\pm0.12$ & $-0.05\color{gray}\pm0.11$ & $-0.04\color{gray}\pm0.12$ & $-0.04\color{gray}\pm0.12$ & $-0.03\color{gray}\pm0.10$ \\
NLL & $\mathbf{0.10}\color{gray}\pm0.07$ & $0.44\color{gray}\pm0.09$ & $0.12\color{gray}\pm0.07$ & $0.91\color{gray}\pm0.74$ & $0.38\color{gray}\pm0.17$ & $0.35\color{gray}\pm0.11$ & $0.19\color{gray}\pm0.14$ & $0.26\color{gray}\pm0.22$ & $0.13\color{gray}\pm0.09$ & $0.44\color{gray}\pm0.04$ & $\mathbf{0.10}\color{gray}\pm0.07$ & $0.11\color{gray}\pm0.07$ & $\mathbf{0.10}\color{gray}\pm0.07$ \\
\hline
\multicolumn{14}{l}{}\\[-2ex]
\multicolumn{14}{l}{* proposed methods}
\end{tabular}
}
\end{table*}

\begin{table*}[p]
\vspace{-20ex}
\centering
\footnotesize
\setlength{\tabcolsep}{1pt}  
\caption{Results for Image Datasets (with the best method for each metric/dataset in \textbf{bold}).}
\label{table:results-images}
\rotatebox{90}{
\begin{tabular}{l|llll|lllll|llll}
\hline\multicolumn{1}{l|}{\bf Dataset} & \multicolumn{4}{c|}{\bf Non-unimodal methods} & \multicolumn{5}{c|}{\bf Unimodal hard constraints methods} & \multicolumn{4}{c}{\bf Unimodal soft constraints methods} \\\hline
\bf BACH & CE & POM & OE & CDW & BU & PU & ORD-ACL & VS-SL & UN* & UR & CO2 & WU-KLDIV* & WU-Wass* \\
\%Accuracy & $86.2\color{gray}\pm6.8$ & $50.3\color{gray}\pm2.1$ & $86.6\color{gray}\pm6.0$ & $78.1\color{gray}\pm14.8$ & $83.8\color{gray}\pm2.0$ & $62.5\color{gray}\pm7.8$ & $80.3\color{gray}\pm11.2$ & $75.6\color{gray}\pm8.7$ & $82.5\color{gray}\pm6.2$ & $85.9\color{gray}\pm3.1$ & $83.8\color{gray}\pm4.7$ & $82.5\color{gray}\pm5.4$ & $\mathbf{89.1}\color{gray}\pm4.1$ \\
MAE & $0.20\color{gray}\pm0.10$ & $0.58\color{gray}\pm0.04$ & $0.17\color{gray}\pm0.08$ & $0.23\color{gray}\pm0.14$ & $0.18\color{gray}\pm0.02$ & $0.43\color{gray}\pm0.11$ & $0.28\color{gray}\pm0.16$ & $0.30\color{gray}\pm0.12$ & $0.25\color{gray}\pm0.10$ & $0.20\color{gray}\pm0.02$ & $0.23\color{gray}\pm0.08$ & $0.24\color{gray}\pm0.07$ & $\mathbf{0.16}\color{gray}\pm0.07$ \\
QWK & $87.4\color{gray}\pm6.5$ & $77.4\color{gray}\pm3.3$ & $90.1\color{gray}\pm5.7$ & $89.1\color{gray}\pm6.3$ & $\mathbf{90.9}\color{gray}\pm1.8$ & $82.7\color{gray}\pm5.9$ & $79.2\color{gray}\pm13.3$ & $83.8\color{gray}\pm7.4$ & $83.8\color{gray}\pm8.4$ & $87.4\color{gray}\pm0.9$ & $84.7\color{gray}\pm5.8$ & $85.8\color{gray}\pm3.4$ & $89.0\color{gray}\pm5.2$ \\
\%$\tau$ & $84.0\color{gray}\pm7.4$ & $72.9\color{gray}\pm2.8$ & $87.0\color{gray}\pm6.9$ & $85.3\color{gray}\pm7.8$ & $87.5\color{gray}\pm1.9$ & $77.9\color{gray}\pm6.7$ & $75.7\color{gray}\pm14.7$ & $78.7\color{gray}\pm8.9$ & $79.5\color{gray}\pm9.6$ & $83.8\color{gray}\pm0.9$ & $80.9\color{gray}\pm6.7$ & $82.0\color{gray}\pm4.4$ & $86.5\color{gray}\pm5.6$ \\
\%Unimodal & $80.0\color{gray}\pm7.6$ & $32.5\color{gray}\pm1.8$ & $81.6\color{gray}\pm4.5$ & $99.1\color{gray}\pm1.2$ & $\mathbf{100}\color{gray}\pm0.0$ & $\mathbf{100}\color{gray}\pm0.0$ & $\mathbf{100}\color{gray}\pm0.0$ & $\mathbf{100}\color{gray}\pm0.0$ & $\mathbf{100}\color{gray}\pm0.0$ & $94.1\color{gray}\pm3.4$ & $79.7\color{gray}\pm7.2$ & $90.9\color{gray}\pm3.1$ & $83.8\color{gray}\pm4.2$ \\
ZME & $-0.02\color{gray}\pm0.08$ & $0.06\color{gray}\pm0.11$ & $-0.01\color{gray}\pm0.07$ & $0.01\color{gray}\pm0.07$ & $-0.01\color{gray}\pm0.07$ & $-0.01\color{gray}\pm0.16$ & $0.06\color{gray}\pm0.05$ & $\mathbf{0.00}\color{gray}\pm0.09$ & $-0.01\color{gray}\pm0.06$ & $\mathbf{0.00}\color{gray}\pm0.05$ & $\mathbf{0.00}\color{gray}\pm0.10$ & $0.06\color{gray}\pm0.09$ & $-0.02\color{gray}\pm0.06$ \\
NLL & $0.53\color{gray}\pm0.28$ & $2.01\color{gray}\pm0.06$ & $\mathbf{0.42}\color{gray}\pm0.19$ & $0.69\color{gray}\pm0.45$ & $0.69\color{gray}\pm0.18$ & $0.87\color{gray}\pm0.17$ & $0.90\color{gray}\pm0.66$ & $0.83\color{gray}\pm0.56$ & $0.61\color{gray}\pm0.23$ & $0.76\color{gray}\pm0.04$ & $0.71\color{gray}\pm0.09$ & $0.56\color{gray}\pm0.17$ & $0.50\color{gray}\pm0.26$ \\
\hline
\bf DHCI & CE & POM & OE & CDW & BU & PU & ORD-ACL & VS-SL & UN* & UR & CO2 & WU-KLDIV* & WU-Wass* \\
\%Accuracy & $53.9\color{gray}\pm2.3$ & $42.9\color{gray}\pm2.1$ & $55.7\color{gray}\pm2.2$ & $37.2\color{gray}\pm5.4$ & $47.5\color{gray}\pm3.5$ & $44.4\color{gray}\pm3.5$ & $49.7\color{gray}\pm5.2$ & $50.0\color{gray}\pm2.3$ & $53.4\color{gray}\pm1.1$ & $\mathbf{57.0}\color{gray}\pm5.1$ & $56.1\color{gray}\pm3.1$ & $56.1\color{gray}\pm2.4$ & $56.2\color{gray}\pm2.3$ \\
MAE & $0.73\color{gray}\pm0.03$ & $0.94\color{gray}\pm0.06$ & $0.67\color{gray}\pm0.05$ & $0.80\color{gray}\pm0.04$ & $0.70\color{gray}\pm0.04$ & $0.87\color{gray}\pm0.08$ & $0.74\color{gray}\pm0.07$ & $0.70\color{gray}\pm0.04$ & $0.70\color{gray}\pm0.02$ & $0.68\color{gray}\pm0.08$ & $\mathbf{0.66}\color{gray}\pm0.06$ & $0.70\color{gray}\pm0.05$ & $0.75\color{gray}\pm0.06$ \\
QWK & $62.8\color{gray}\pm2.4$ & $62.2\color{gray}\pm4.1$ & $67.9\color{gray}\pm2.8$ & $59.9\color{gray}\pm1.2$ & $\mathbf{69.4}\color{gray}\pm2.7$ & $62.2\color{gray}\pm5.8$ & $62.8\color{gray}\pm5.6$ & $66.6\color{gray}\pm3.0$ & $66.3\color{gray}\pm1.2$ & $65.7\color{gray}\pm6.4$ & $67.0\color{gray}\pm4.5$ & $65.4\color{gray}\pm3.4$ & $61.7\color{gray}\pm3.9$ \\
\%$\tau$ & $55.8\color{gray}\pm1.4$ & $55.9\color{gray}\pm3.6$ & $59.2\color{gray}\pm3.5$ & $54.7\color{gray}\pm2.9$ & $59.9\color{gray}\pm2.8$ & $55.9\color{gray}\pm3.8$ & $54.4\color{gray}\pm4.9$ & $57.7\color{gray}\pm3.3$ & $59.1\color{gray}\pm1.1$ & $59.2\color{gray}\pm5.2$ & $60.3\color{gray}\pm4.3$ & $58.7\color{gray}\pm3.5$ & $55.3\color{gray}\pm4.3$ \\
\%Unimodal & $59.2\color{gray}\pm6.1$ & $13.1\color{gray}\pm1.9$ & $61.6\color{gray}\pm2.6$ & $\mathbf{100}\color{gray}\pm0.0$ & $\mathbf{100}\color{gray}\pm0.0$ & $\mathbf{100}\color{gray}\pm0.0$ & $\mathbf{100}\color{gray}\pm0.0$ & $\mathbf{100}\color{gray}\pm0.0$ & $\mathbf{100}\color{gray}\pm0.0$ & $80.0\color{gray}\pm4.4$ & $93.0\color{gray}\pm1.2$ & $57.4\color{gray}\pm2.7$ & $51.5\color{gray}\pm3.0$ \\
ZME & $-0.20\color{gray}\pm0.15$ & $\mathbf{-0.03}\color{gray}\pm0.20$ & $-0.06\color{gray}\pm0.05$ & $\mathbf{-0.03}\color{gray}\pm0.11$ & $-0.08\color{gray}\pm0.07$ & $-0.21\color{gray}\pm0.18$ & $\mathbf{-0.03}\color{gray}\pm0.12$ & $-0.05\color{gray}\pm0.16$ & $-0.18\color{gray}\pm0.02$ & $-0.19\color{gray}\pm0.13$ & $-0.20\color{gray}\pm0.07$ & $-0.19\color{gray}\pm0.08$ & $-0.11\color{gray}\pm0.16$ \\
NLL & $2.27\color{gray}\pm0.28$ & $2.24\color{gray}\pm0.04$ & $1.54\color{gray}\pm0.08$ & $3.47\color{gray}\pm1.08$ & $1.27\color{gray}\pm0.10$ & $1.25\color{gray}\pm0.07$ & $2.05\color{gray}\pm0.22$ & $2.34\color{gray}\pm0.11$ & $1.79\color{gray}\pm0.12$ & $\mathbf{1.17}\color{gray}\pm0.02$ & $1.36\color{gray}\pm0.08$ & $1.93\color{gray}\pm0.12$ & $2.05\color{gray}\pm0.13$ \\
\hline
\bf FGNET & CE & POM & OE & CDW & BU & PU & ORD-ACL & VS-SL & UN* & UR & CO2 & WU-KLDIV* & WU-Wass* \\
\%Accuracy & $10.4\color{gray}\pm0.7$ & $4.4\color{gray}\pm0.2$ & $15.0\color{gray}\pm1.5$ & $6.1\color{gray}\pm2.0$ & $11.9\color{gray}\pm3.3$ & $\mathbf{16.6}\color{gray}\pm4.4$ & $13.9\color{gray}\pm3.4$ & $13.1\color{gray}\pm3.4$ & $12.4\color{gray}\pm3.0$ & $12.1\color{gray}\pm2.2$ & $10.7\color{gray}\pm1.7$ & $12.0\color{gray}\pm2.9$ & $11.2\color{gray}\pm2.6$ \\
MAE & $5.93\color{gray}\pm0.46$ & $13.30\color{gray}\pm0.19$ & $\mathbf{3.29}\color{gray}\pm0.27$ & $5.91\color{gray}\pm1.15$ & $3.63\color{gray}\pm0.50$ & $3.40\color{gray}\pm0.15$ & $3.53\color{gray}\pm0.14$ & $3.76\color{gray}\pm0.46$ & $3.40\color{gray}\pm0.24$ & $4.07\color{gray}\pm0.16$ & $3.90\color{gray}\pm0.37$ & $4.07\color{gray}\pm0.53$ & $3.74\color{gray}\pm0.35$ \\
QWK & $67.5\color{gray}\pm8.4$ & $47.7\color{gray}\pm5.2$ & $\mathbf{92.0}\color{gray}\pm2.0$ & $77.9\color{gray}\pm9.1$ & $91.7\color{gray}\pm1.8$ & $\mathbf{92.0}\color{gray}\pm1.0$ & $91.1\color{gray}\pm0.9$ & $89.1\color{gray}\pm3.1$ & $91.4\color{gray}\pm1.8$ & $87.4\color{gray}\pm1.2$ & $85.6\color{gray}\pm5.7$ & $86.9\color{gray}\pm2.7$ & $89.0\color{gray}\pm3.5$ \\
\%$\tau$ & $61.2\color{gray}\pm3.4$ & $42.5\color{gray}\pm0.8$ & $81.1\color{gray}\pm1.3$ & $65.9\color{gray}\pm6.7$ & $79.7\color{gray}\pm2.6$ & $82.4\color{gray}\pm0.7$ & $79.9\color{gray}\pm0.9$ & $79.2\color{gray}\pm2.2$ & $80.7\color{gray}\pm1.5$ & $74.5\color{gray}\pm2.6$ & $77.6\color{gray}\pm2.2$ & $75.0\color{gray}\pm3.7$ & $77.7\color{gray}\pm0.5$ \\
\%Unimodal & $3.4\color{gray}\pm1.5$ & $2.9\color{gray}\pm1.1$ & $11.4\color{gray}\pm3.4$ & $15.7\color{gray}\pm14.1$ & $\mathbf{100}\color{gray}\pm0.0$ & $\mathbf{100}\color{gray}\pm0.0$ & $\mathbf{100}\color{gray}\pm0.0$ & $\mathbf{100}\color{gray}\pm0.0$ & $\mathbf{100}\color{gray}\pm0.0$ & $26.0\color{gray}\pm1.2$ & $67.4\color{gray}\pm2.4$ & $19.7\color{gray}\pm3.0$ & $57.7\color{gray}\pm4.0$ \\
ZME & $-2.49\color{gray}\pm1.21$ & $-11.73\color{gray}\pm0.76$ & $-0.09\color{gray}\pm0.41$ & $1.16\color{gray}\pm1.30$ & $0.25\color{gray}\pm1.12$ & $\mathbf{-0.03}\color{gray}\pm1.18$ & $-0.24\color{gray}\pm0.66$ & $-0.62\color{gray}\pm0.46$ & $-0.09\color{gray}\pm0.36$ & $-0.91\color{gray}\pm0.40$ & $-1.32\color{gray}\pm0.87$ & $-1.17\color{gray}\pm0.11$ & $-1.50\color{gray}\pm0.83$ \\
NLL & $4.31\color{gray}\pm0.22$ & $7.07\color{gray}\pm0.04$ & $2.92\color{gray}\pm0.10$ & $9.22\color{gray}\pm0.59$ & $3.02\color{gray}\pm0.24$ & $\mathbf{2.73}\color{gray}\pm0.04$ & $3.83\color{gray}\pm0.14$ & $3.49\color{gray}\pm0.18$ & $3.28\color{gray}\pm0.06$ & $2.94\color{gray}\pm0.05$ & $2.92\color{gray}\pm0.09$ & $2.96\color{gray}\pm0.12$ & $2.81\color{gray}\pm0.09$ \\
\hline
\bf Pap Smear & CE & POM & OE & CDW & BU & PU & ORD-ACL & VS-SL & UN* & UR & CO2 & WU-KLDIV* & WU-Wass* \\
\%Accuracy & $78.1\color{gray}\pm4.5$ & $43.9\color{gray}\pm4.1$ & $78.7\color{gray}\pm4.8$ & $73.9\color{gray}\pm2.6$ & $71.7\color{gray}\pm5.0$ & $49.8\color{gray}\pm1.9$ & $77.9\color{gray}\pm2.2$ & $62.7\color{gray}\pm8.3$ & $80.7\color{gray}\pm4.9$ & $\mathbf{81.1}\color{gray}\pm2.7$ & $76.8\color{gray}\pm2.5$ & $78.7\color{gray}\pm5.8$ & $79.2\color{gray}\pm4.1$ \\
MAE & $0.27\color{gray}\pm0.10$ & $0.67\color{gray}\pm0.08$ & $0.23\color{gray}\pm0.06$ & $0.27\color{gray}\pm0.03$ & $0.29\color{gray}\pm0.06$ & $0.53\color{gray}\pm0.02$ & $0.26\color{gray}\pm0.04$ & $0.39\color{gray}\pm0.09$ & $0.22\color{gray}\pm0.06$ & $\mathbf{0.20}\color{gray}\pm0.03$ & $0.27\color{gray}\pm0.03$ & $0.24\color{gray}\pm0.07$ & $0.23\color{gray}\pm0.05$ \\
QWK & $89.5\color{gray}\pm6.6$ & $82.3\color{gray}\pm3.0$ & $92.6\color{gray}\pm2.6$ & $91.1\color{gray}\pm1.4$ & $91.5\color{gray}\pm1.8$ & $87.5\color{gray}\pm0.3$ & $90.7\color{gray}\pm2.5$ & $89.7\color{gray}\pm1.9$ & $92.3\color{gray}\pm3.0$ & $\mathbf{93.3}\color{gray}\pm1.2$ & $89.9\color{gray}\pm2.0$ & $91.4\color{gray}\pm2.9$ & $92.2\color{gray}\pm1.8$ \\
\%$\tau$ & $80.5\color{gray}\pm7.4$ & $76.0\color{gray}\pm4.2$ & $84.6\color{gray}\pm3.7$ & $81.1\color{gray}\pm2.2$ & $81.3\color{gray}\pm3.2$ & $80.5\color{gray}\pm0.9$ & $80.8\color{gray}\pm2.9$ & $80.1\color{gray}\pm3.4$ & $84.0\color{gray}\pm5.8$ & $85.2\color{gray}\pm2.4$ & $80.3\color{gray}\pm2.0$ & $82.4\color{gray}\pm5.2$ & $84.1\color{gray}\pm4.2$ \\
\%Unimodal & $82.2\color{gray}\pm2.9$ & $16.4\color{gray}\pm4.1$ & $89.3\color{gray}\pm5.6$ & $\mathbf{100}\color{gray}\pm0.0$ & $\mathbf{100}\color{gray}\pm0.0$ & $\mathbf{100}\color{gray}\pm0.0$ & $\mathbf{100}\color{gray}\pm0.0$ & $\mathbf{100}\color{gray}\pm0.0$ & $\mathbf{100}\color{gray}\pm0.0$ & $98.0\color{gray}\pm1.8$ & $84.4\color{gray}\pm5.7$ & $85.5\color{gray}\pm3.5$ & $95.6\color{gray}\pm3.1$ \\
ZME & $-0.08\color{gray}\pm0.13$ & $0.34\color{gray}\pm0.10$ & $-0.04\color{gray}\pm0.12$ & $-0.02\color{gray}\pm0.06$ & $0.08\color{gray}\pm0.07$ & $0.28\color{gray}\pm0.03$ & $0.05\color{gray}\pm0.08$ & $-0.06\color{gray}\pm0.15$ & $\mathbf{0.01}\color{gray}\pm0.08$ & $-0.02\color{gray}\pm0.06$ & $\mathbf{-0.01}\color{gray}\pm0.09$ & $-0.06\color{gray}\pm0.05$ & $-0.07\color{gray}\pm0.07$ \\
NLL & $0.88\color{gray}\pm0.23$ & $2.75\color{gray}\pm0.06$ & $0.80\color{gray}\pm0.34$ & $0.72\color{gray}\pm0.17$ & $0.86\color{gray}\pm0.09$ & $0.95\color{gray}\pm0.03$ & $0.91\color{gray}\pm0.08$ & $1.06\color{gray}\pm0.28$ & $0.81\color{gray}\pm0.15$ & $0.76\color{gray}\pm0.01$ & $0.94\color{gray}\pm0.15$ & $0.81\color{gray}\pm0.15$ & $\mathbf{0.58}\color{gray}\pm0.05$ \\
\hline
\bf FocusPath & CE & POM & OE & CDW & BU & PU & ORD-ACL & VS-SL & UN* & UR & CO2 & WU-KLDIV* & WU-Wass* \\
\%Accuracy & $77.4\color{gray}\pm1.3$ & $49.8\color{gray}\pm3.6$ & $80.1\color{gray}\pm1.2$ & $47.5\color{gray}\pm11.8$ & $65.3\color{gray}\pm6.4$ & $28.0\color{gray}\pm2.8$ & $78.5\color{gray}\pm2.0$ & $80.8\color{gray}\pm2.9$ & $80.6\color{gray}\pm5.0$ & $\mathbf{81.9}\color{gray}\pm1.3$ & $78.4\color{gray}\pm3.4$ & $77.9\color{gray}\pm4.2$ & $74.5\color{gray}\pm3.1$ \\
MAE & $0.24\color{gray}\pm0.01$ & $0.61\color{gray}\pm0.04$ & $0.20\color{gray}\pm0.01$ & $0.63\color{gray}\pm0.20$ & $0.36\color{gray}\pm0.07$ & $0.76\color{gray}\pm0.03$ & $0.22\color{gray}\pm0.02$ & $0.20\color{gray}\pm0.03$ & $0.20\color{gray}\pm0.05$ & $\mathbf{0.19}\color{gray}\pm0.01$ & $0.22\color{gray}\pm0.04$ & $0.23\color{gray}\pm0.04$ & $0.27\color{gray}\pm0.03$ \\
QWK & $97.6\color{gray}\pm0.2$ & $94.2\color{gray}\pm0.4$ & $98.3\color{gray}\pm0.2$ & $92.5\color{gray}\pm3.7$ & $97.3\color{gray}\pm0.5$ & $94.5\color{gray}\pm0.3$ & $98.1\color{gray}\pm0.2$ & $\mathbf{98.4}\color{gray}\pm0.2$ & $\mathbf{98.4}\color{gray}\pm0.4$ & $\mathbf{98.4}\color{gray}\pm0.2$ & $98.1\color{gray}\pm0.4$ & $97.9\color{gray}\pm0.4$ & $97.6\color{gray}\pm0.3$ \\
\%$\tau$ & $94.7\color{gray}\pm0.3$ & $94.0\color{gray}\pm0.4$ & $95.9\color{gray}\pm0.2$ & $86.0\color{gray}\pm5.9$ & $94.5\color{gray}\pm0.6$ & $95.5\color{gray}\pm0.3$ & $95.3\color{gray}\pm0.5$ & $95.9\color{gray}\pm0.5$ & $96.0\color{gray}\pm1.0$ & $96.3\color{gray}\pm0.4$ & $95.7\color{gray}\pm0.5$ & $95.0\color{gray}\pm0.8$ & $94.3\color{gray}\pm0.6$ \\
\%Unimodal & $96.9\color{gray}\pm0.5$ & $1.7\color{gray}\pm0.9$ & $98.7\color{gray}\pm0.2$ & $95.2\color{gray}\pm9.6$ & $\mathbf{100}\color{gray}\pm0.0$ & $\mathbf{100}\color{gray}\pm0.0$ & $\mathbf{100}\color{gray}\pm0.0$ & $\mathbf{100}\color{gray}\pm0.0$ & $\mathbf{100}\color{gray}\pm0.0$ & $99.7\color{gray}\pm0.1$ & $99.9\color{gray}\pm0.0$ & $97.1\color{gray}\pm0.2$ & $98.2\color{gray}\pm0.3$ \\
ZME & $-0.05\color{gray}\pm0.06$ & $-0.50\color{gray}\pm0.07$ & $-0.11\color{gray}\pm0.01$ & $-0.06\color{gray}\pm0.19$ & $-0.30\color{gray}\pm0.08$ & $-0.63\color{gray}\pm0.06$ & $\mathbf{-0.02}\color{gray}\pm0.07$ & $-0.07\color{gray}\pm0.06$ & $-0.10\color{gray}\pm0.06$ & $-0.10\color{gray}\pm0.02$ & $-0.11\color{gray}\pm0.06$ & $-0.08\color{gray}\pm0.04$ & $-0.13\color{gray}\pm0.03$ \\
NLL & $0.74\color{gray}\pm0.01$ & $2.39\color{gray}\pm0.03$ & $0.57\color{gray}\pm0.04$ & $2.15\color{gray}\pm0.98$ & $1.34\color{gray}\pm0.02$ & $1.63\color{gray}\pm0.01$ & $0.60\color{gray}\pm0.05$ & $\mathbf{0.55}\color{gray}\pm0.13$ & $0.57\color{gray}\pm0.14$ & $0.84\color{gray}\pm0.02$ & $1.47\color{gray}\pm0.03$ & $0.70\color{gray}\pm0.13$ & $0.75\color{gray}\pm0.10$ \\
\hline
\multicolumn{14}{l}{}\\[-2ex]
\multicolumn{14}{l}{* proposed methods}
\end{tabular}
}
\end{table*}

\input{imgs/outputs}

Results for the various methods are shown in Table~\ref{table:results-tabular} for tabular datasets (using a multi-layer perceptron with one hidden-layer) and Table~\ref{table:results-images} for the images datasets (using a ResNet-18). The methods are divided into three groups: the first group is non-unimodal strategies, the second group is the hard unimodal models, and finally, the soft unimodal losses. To make results easier to read and to measure consistency, Table~\ref{table:results-average} summarizes results by averaging the previous results. In addition, results are also shown only for datasets with IR\textgreater1.5, which is a common threshold to distinguish between low and high imbalance datasets~\cite{perez2014graph}.

On average, across all datasets, the methods have obtained an accuracy of 62.8\%, with a maximum of 68.0\% (OE). In general, there are no big discrepancies, and our proposed methods (UN*, WU-KLDiv*, and WU-Wass*) have reached close to the maximum, 66--67\%, with the advantage of the output probabilities being more unimodal. The results for the other ordinal metrics (QWK, Kendall's $\tau$, and MAE) are similar to Accuracy with a slight advantage to the ordinal methods; in particular, the proposed WU methods show better performance than the other soft unimodal methods for these ordinal metrics. In all cases, the proposed UnimodalNet (UN*) was very close to Ordinal Encoding (OE) with the advantage of having consistently unimodal output probabilities -- OE averaged 78\% on \%Unimodality, while UN* scored 100\%, as would be expected.
Results are also consistent in the low IR or high IR regimes, with the proposed methods performing a little better. However, all methods focus on promoting ordinality, not necessarily on counter-weighting the imbalances between classes.
When considering the bias of the models (ZME), unimodal hard models have slightly less bias than unimodal soft models, which themselves have less bias than non-ordinal models.

In terms of inference time, the methods are all equivalent -- notice that the model is exactly the same when using CrossEntropy (CE) or WU-KLDiv*/WU-Wass*, since only a regularization is added during training. For UN*, a few additions and minimum operations are added (see Fig.~\ref{unimodalNet}), which have a negligible impact on inference time. When it comes to training time, the regularization that is added to WU-KLDiv*/WU-Wass* requires projection of the probabilities that is performed by linear programming (\texttt{scipy.optimize.linprog}, in our case), which makes training time 28\% times slower (on median) than CE.

Why would the soft constraints ordinal methods sometimes surpass the hard constraints ordinal methods? Unlike the proposed method, Binomial (BU) and Poisson (PU) force the output to follow those parametric distributions; therefore, there is an underlying assumption that might be unwarranted and could lead to underfitting. The proposed hard constraint method (UN*) is free from this assumption, yet it still assumes a unimodal output, and it might be the case for some datasets that the ordinality property is a little fuzzy, therefore, some transgression to ordinal consistency might improve performance. Furthermore, the soft methods are based on regularization, which makes them easier to deploy into any variety of losses.

An illustration of the output of each model is exemplified by Fig.~\ref{fig:outputs}. All the hard-constraints models maintain an ordinally-consistent output, while our two soft-constraints methods (WU-KLDiv* and WU-Wass*) tend to nudge the output into the unimodal space.

Since WU-KLDiv is the sum of CE and the proposed regularization, weighted by $\lambda$, an additional study 
illustrating the effects of the $\lambda$ coefficient is shown in Fig.~\ref{figure:lambdas}. As $\lambda$ is increased, and Wasserstein Unimodality is exerted, the trend is towards higher unimodal probability outputs. The drop in accuracy is only slight and even non-existent in some datasets.

A repository containing the proposed ordinal methods and the baselines is available at \url{https://github.com/rpmcruz/unimodal-ordinal-regression}.

\section{Conclusion}
A vast literature on ordinal losses exists. Beyond promoting ordinality, many losses focus on having the model probability outputs follow a unimodal distribution, a property expected by an ordinal model. This unimodality may be either forced into the model (hard-unimodality) or promoted through regularization (soft-unimodality).

The paper identifies and addresses two missing parts of the literature. A novel hard-unimodality method that is not parametric is proposed: UnimodalNet is an activation function that ensures that the architecture outputs are unimodal.

Furthermore, existing soft-unimodal proposals are only heuristically motivated, without a clear mathematical foundation. For that purpose, a regularization term is proposed that identifies the closest unimodal distribution to the output produced by the model through the Wasserstein Distance. It is shown that this guides the optimizer toward a unimodal solution.

In the end, experiments are performed using ten datasets, with UnimodalNet consistently performing second-best on both accuracy and MAE while always ensuring unimodality. The proposed regularization terms are also fairly competitive while providing a mathematical foundation and empirically offering high levels of unimodality.

\bibliographystyle{unsrt} 
\bibliography{paper-arxiv}
\end{document}